\documentclass[lettersize,journal]{IEEEtran}
\usepackage{algorithm}
\usepackage[caption=false,font=normalsize,labelfont=sf,textfont=sf]{subfig}
\usepackage{stfloats}
\usepackage{cite}
\usepackage{amsmath,amssymb,amsfonts}
\usepackage{algorithmic}
\usepackage{graphicx}
\usepackage{textcomp}
\usepackage{multirow}
\usepackage{url}
\usepackage[switch]{lineno}
\usepackage{verbatim}

\usepackage{array}
\usepackage{booktabs}
\usepackage{makecell}
\usepackage{color}
\usepackage{lettrine}

\usepackage{pgfplots}
\usepackage{pgfplotstable}
\pgfplotsset{compat=1.18}
\usepackage{colortbl}
\usepackage{xcolor}
\usepackage{bm}

\usepackage{diagbox}
\usepackage{ragged2e}
\usepackage{threeparttable}

\begin{document}

\title{Coherence-Guided Pyramid Self-Contrastive Disentanglement for Adaptive IQ Signal Denoising in Synthetic Aperture Ultrasound}

\author{Jiajing Zhang, Bingze Dai, Xi Zhang, Yue Xu, Wei-Ning Lee, \textit{Senior Member, IEEE}
\thanks{This work was supported by the Hong Kong Research Grants Council General Research Fund (17205022). Corresponding author: Wei-Ning Lee.}
\thanks{Jiajing Zhang, Bingze Dai, Xi Zhang, and Wei-Ning Lee are with the Department of Electrical and Computer Engineering, The University of Hong Kong, Hong Kong. Wei-Ning Lee is also with the School of Biomedical Engineering, The University of Hong Kong.}
\thanks{Yue Xu was with the Department of Electrical and Computer Engineering, The University of Hong Kong and is currently with the Department of Biomedical Engineering, Duke University, North Carolina, United States.}
}

\markboth{Journal of \LaTeX\ Class Files,~Vol.~14, No.~8, August~2026}
{Zhang \MakeLowercase{\textit{et al.}}: Coherence-Guided Pyramid Self-Contrastive Disentanglement for Adaptive IQ Signal Denoising in Synthetic Aperture Ultrasound}


\maketitle

\begin{abstract}
Synthetic aperture ultrasound (SAU) groups neighboring elements in an array into sub-apertures and sequentially transmits them to achieve two-way focusing by compounding all received signals. However, the inherent electronic noise, side lobes, and speckle noise still complicate clinical interpretation of SAU images. Previous denoising methods either rely on explicit noise assumptions or are pretrained in a limited image domain, thus facing inevitable domain shift under complex \textit{in vivo} noise conditions. This study proposes an adaptive IQ signal denoising method to suppress aperture-dependent degradation before B-mode image formation. We model sub-aperture events in each SAU acquisition as multiple noisy measurements and extend cross-aperture coherence as a contrastive learning critic in a Pyramid Self-Contrastive Disentanglement (PSCD) framework. PSCD encodes the raw IQ tensor into separated coherent and incoherent latent spaces, then reconstructs clean IQ signals exclusively from the former to retain coherent tissue information while suppressing incoherent noise. An Aperture-to-Aperture proxy task self-supervises the reconstruction in dual domains. The whole framework can be optimized at test time using single-shot data without pretraining or clean labels. Simulations including noise levels from 0 to 30 dB demonstrated improvements of 69.3$\%$ in SNR and 34.4$\%$ in CNR. \textit{In vivo} experiments, including six echocardiographic views and abdominal images, showed 84.8$\%$ SNR and 25.7$\%$ CNR gains using only two sub-apertures. Our PSCD also achieved higher image resolution and boundary sharpness, while preserving high-frequency details. These results support adaptive IQ signal denoising leveraging cross-aperture measurement redundancy, enabling precise anatomical visualization without domain shift and pretraining costs.
\end{abstract}


\section{Introduction}
 
\begin{figure}
\centering
\includegraphics[width=\linewidth]{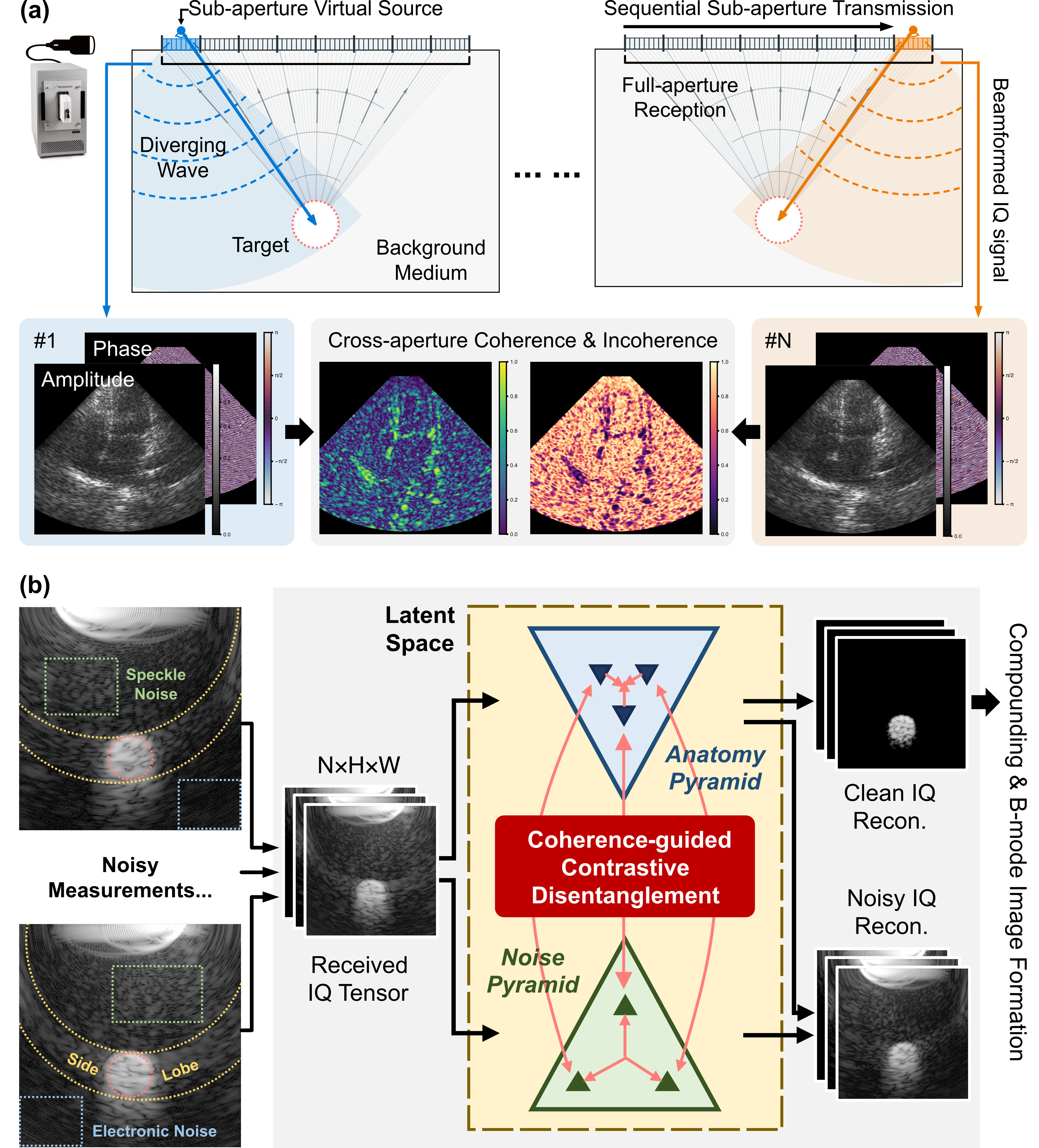}
\caption{\color{black} (a) Top: Synthetic aperture ultrasound imaging using sub-aperture transmission, where each received signal is noisy measurement of the target depended by its virtual source location. Bottom: Examples of beamformed IQ signals measured by different sub-aperture events and their cross-aperture coherence and incoherence. (b) Our PSCD disentangles the IQ tensor of single-shot SAU imaging into a coherent anatomy space and an incoherent noise space, then reconstructs the former as clean IQ signals for B-mode image formation.}
\label{solution}
\end{figure}

\lettrine{U}{ltrasound} is an indispensable imaging modality in modern healthcare systems, offering unique advantages for dynamic and real-time assessment without ionizing radiation. Conventional ultrasound imaging is based on single transmit focusing, so optimal spatial resolution is limited to the pre-defined foci. As shown in Fig. \ref{solution}(a, top), achieving two-way focus throughout the field of view, synthetic aperture ultrasound (SAU) divides an array transducer into multiple sub-apertures and synthesizes transmit focusing by coherently compounding beamformed signals from individual sub-apertures \cite{jensen2006synthetic}. However, at high-frame-rate imaging using unfocused wave transmission (e.g. diverging wave), low signal-to-noise ratio (SNR) due to strong noise still hampers its precise clinical interpretation. 

The left part of Fig. \ref{solution}(b) illustrates major noise sources in SAU imaging, including side lobes, electronic noise, and speckle noise. Electronic noise primarily arises from electromagnetic interference and electronics. As grainy patterns (green boxes in Fig. \ref{solution}(b)), speckle noise results from wave interference among echoes scattered by sub-wavelength structures \cite{krissian2005speckle}; Side lobes stem from low-energy and off-axis echoes and may mask hypoechoic structures. These noises degrade image quality by obscuring anatomical boundaries, distorting fine structures, and reducing image contrast.

\begin{figure}
\centering
\includegraphics[width=\linewidth]{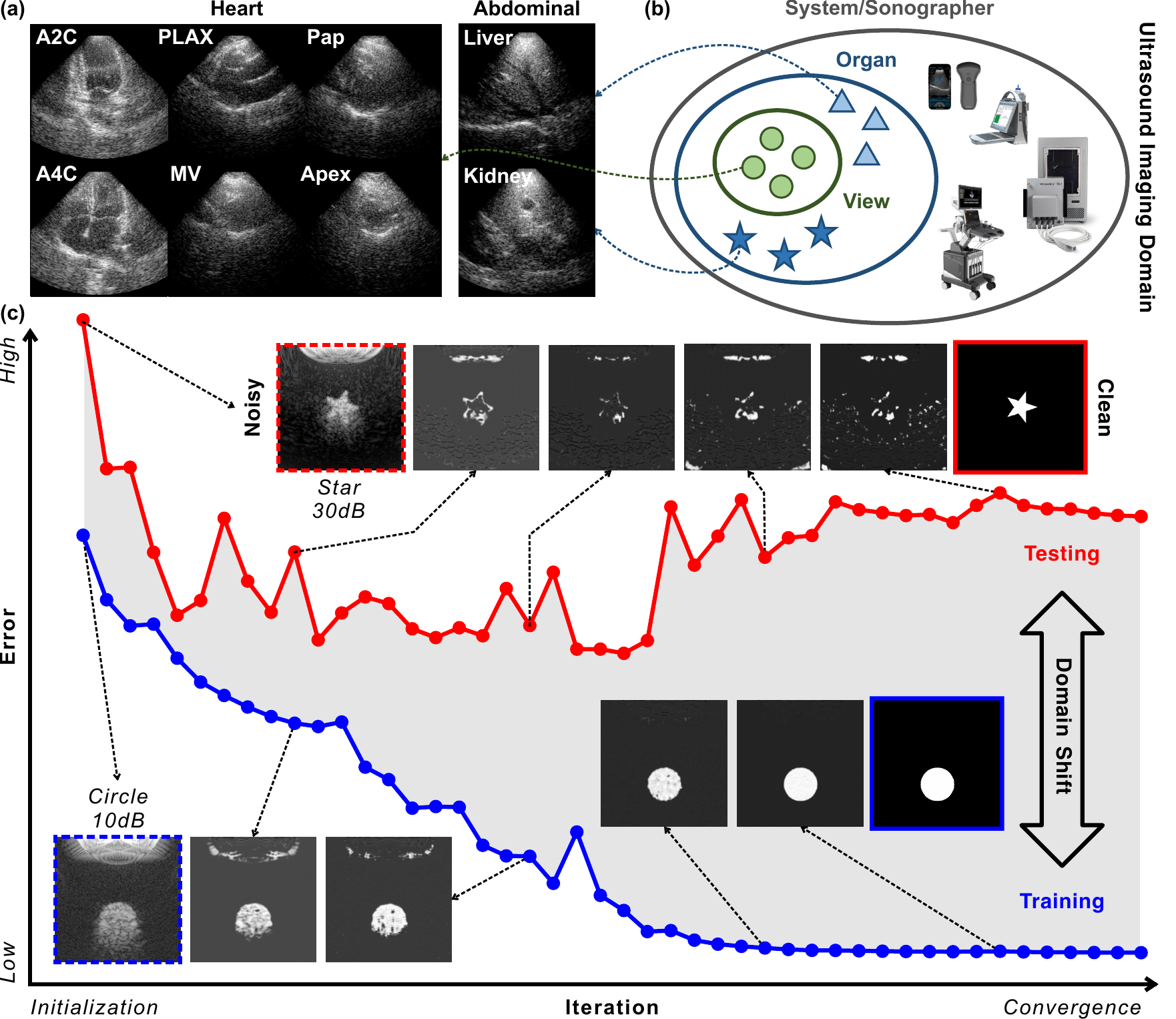}
\caption{(a) Examples of \textit{in vivo} ultrasound images in (b) various imaging domains. (c) Illustration of domain shift under different SNRs and inclusion geometries for training (blue) and testing (red).}
\label{domain}
\end{figure}

Despite extensive efforts in ultrasound denoising (Section \ref{related}), domain shift is still one of the most critical challenges. Conventional methods based on filters \cite{gupta2018performance, dabov2007image}, similarity \cite{yu2002speckle, yang2016local, li2025research, gan2015bm3d}, and coherence factors \cite{hollman1999coherence, li2003adaptive, camacho2009phase, wang2007performance, asl2009minimum} rely on explicit noise models that may not generalize across complex and dynamic imaging conditions. In practice, electronic noise varies between systems with different driving voltages and transmission frequencies, yielding different SNR levels. Speckle noise differs across organs because tissue-dependent scattering precludes a universal probabilistic representation \cite{li2025speckle2self}. Furthermore, wave attenuation and time-gain compensation (TGC) cause depth-dependent noise levels. As illustrated in Fig. \ref{domain}(a), these factors produce considerable variations across organs and scan views. Therefore, model- or coherence-based methods \cite{long2022spatial, lediju2011short, wang2017short} can hardly adapt to diverse imaging domains with sensitive parameters, and images tend to be over-smoothed under domain shift.

Recently, learning-based methods have employed regression models to learn implicit noise representations. Such data-driven approaches address the problem of unknown noise priors, but at the expense of massive labeled data requirements and a costly pretraining process. Their training is usually confined to a fixed data domain. For ultrasound imaging, a modality prized for its flexibility, a system must handle diverse organs, scan views, and even sonographer preferences. All these factors imply significant domain shifts (Fig. \ref{domain}(b)) when testing these methods on real-world clinical environments.

\begin{figure*}
\centering
\includegraphics[width=\textwidth]{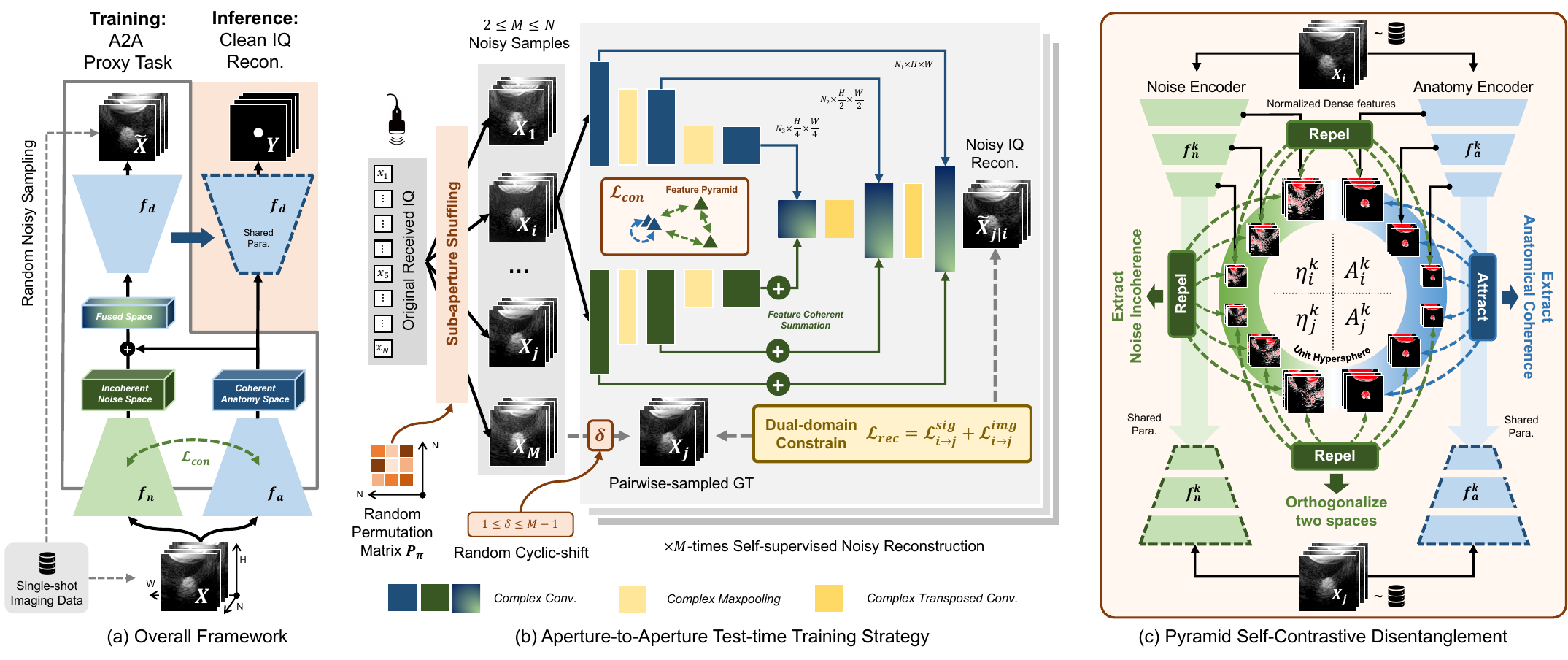}
\caption{(a) The overall framework: Given IQ data $X$ of single-shot SAU imaging, an anatomy encoder $f_a$ and a noise encoder $f_n$ encode it into coherent anatomy space and incoherent noise space, respectively. A decoder $f_d$ reconstructs another noisy $\tilde{X}$ from the fused space as a self-supervised proxy task for model optimization. After training, $f_d$ decodes the anatomy space exclusively to infer the clean IQ reconstruction. (b) A2A strategy for test-time training: $M$ noisy samples are first generated by random sub-aperture shuffling. Each noisy sample (e.g., $X_i$) performs a directional noisy reconstruction to produce $X_{j|i}$. Loss $\mathcal{L}_{\mathrm{rec}}$ samples another noisy $X_j$ to constrain $X_{j|i}$ in both the IQ signal and B-mode image domains. (c) Pyramid self-contrastive disentanglement: For any unordered pair $(X_i, X_j)$, $\{\mathcal{A}_{i}^k, \mathcal{A}_{j}^k, \eta_{i}^k, \eta_{j}^k | k=1,...,K\}$ construct anatomy and noise feature pyramids from $f_a$ and $f_n$. Loss $\mathcal{L}_{\mathrm{con}}$ attracts $(\mathcal{A}_i^k, \mathcal{A}_j^k)$ to extract coherence among $(X_i, X_j)$ into anatomy space, repels $(\eta_i^k, \eta_j^k)$ to extract incoherent features into the noise space, and orthogonalizes the two spaces by repelling $(\mathcal{A}_{i/j}^k, \eta_{i/j}^k)$ to reduce feature redundancy. The complementarity of these two spaces to represent the complete IQ tensor is ensured by the A2A reconstruction.}
\label{model}
\end{figure*}

Ultrasound noise is coupled with anatomical structures. Regression models learn a direct mapping from noisy to clean images rather than the underlying noise characteristics. Pretrained models are favorable in the training domain, but they can degrade performance or even corrupt unseen anatomy and introduce artifacts when applied to unpredictable testing domains. This phenomenon is illustrated in Fig. \ref{domain}(c). A denoising model trained until convergence on a dataset of circular inclusion images with an SNR of 10 dB failed on a star inclusion image with an SNR of 30 dB and severely broke the inclusion boundary. Ultrasound denoising methods generalizable to various imaging domains remain one of the important challenges. Exploiting information in the raw IQ signals measured within individual SAU acquisition offers a promising solution to adaptive denoising before B-mode image formation.

Data scarcity constitutes another challenge. In dynamic imaging applications, such as \textit{in vivo} echocardiography, acquiring registered clean and noisy data pairs of a beating heart is physically infeasible. Furthermore, existing public datasets and pre-trained models predominantly operate on beamformed B-mode images. This confines current methodologies to the post-processing paradigm, precluding the effective utilization of raw IQ signals before final B-mode image formation. A significant research gap persists in image denoising by exploiting the data redundancy in IQ measurements acquired across sub-aperture transmit–receive events in SAU imaging.

To tackle the aforementioned challenges, this paper proposes a coherence-guided IQ signal denoising framework, named Pyramid Self-Contrastive Disentanglement (PSCD), to adaptively reconstruct the underlying anatomical information in single-shot SAU imaging data. PSCD operates on complex IQ signals before compounding and B-mode image formation. This work provides the following contributions:

\begin{itemize}
    \item \textbf{PSCD} disentangles the anatomical coherence and noise incoherence among multiple noisy IQ measurements into two pyramid latent spaces by generalizing the cross-aperture coherence prior into a contrastive learning objective;
    \item \textbf{A2A} (Aperture-to-Aperture) forms a self-supervised proxy task for pure test-time training using only single-shot SAU imaging data by dual-domain approximation among random noisy samples with shuffled sub-apertures;
    \item \textbf{\textit{In vivo} and simulation evaluations} show that this framework enhances SNR, image contrast, and resolution while eliminating the domain shift, pretraining costs, and labeled dataset reliance.
\end{itemize}

\section{Related Works}
\label{related}

\subsection{Model-based Ultrasound Denoising}
Ultrasound image denoising dates back to the 1970s \cite{kuc1979application} when global filters were integrated into systems. Adaptive filters based on local noise statistics were subsequently introduced, including Lee, Wiener, and Frost filters, etc \cite{gupta2018performance}. Speckle reducing anisotropic diffusion (SRAD) \cite{yu2002speckle} applied anisotropic diffusion to despeckle. Wavelet-based methods combine the advantages of global and local methods \cite{rizi2011wavelet, andria2013suitable}. Non-local methods such as BM3D (block-matching and 3D filtering) \cite{gan2015bm3d} and WNNM (weighted nuclear norm minimization) \cite{khare2021speckle} were adapted for ultrasound images \cite{yang2016local, li2025research} under the low-rank prior \cite{zhu2017non}. These methods are derived from explicit noise assumptions and are usually sensitive to parameters such as window size and threshold.

\subsection{Coherence-based Ultrasound Denoising}
\label{related_coherence}

Tissue inhomogeneities can cause phase distortion, resulting in focusing error and an increased level of side lobes. The coherence factor (CF) \cite{hollman1999coherence}, which quantifies the coherent and incoherent signal energy components as a focusing quality factor, has been devised to weight aperture data to reduce the side lobes \cite{hollman1999coherence}. Based upon similar principles, the phase coherence factor (PCF) \cite{camacho2009phase} and other variants \cite{li2003adaptive, wang2007performance, asl2009minimum} were proposed. Alternatively, short-lag spatial coherence imaging \cite{lediju2011short} visualized the coherence values as the final images \cite{bell2013short, wang2017short}. Incoherent spatial compounding has been established for electronic and speckle noise suppression \cite{li1994evaluational}. Related covariance-based adaptive beamforming also exploits correlations across multi-angle plane-wave measurements \cite{pu2024minimum}. These methods leverage coherence across receive channels or signal diversity across acquisitions to suppress noise, extending beyond noise models defined in B-mode images \cite{Long_Trahey_Bottenus_2022}.

\subsection{Learning-based Ultrasound Denoising}

Learning-based methods formulate denoising as an image-to-image regression using RNN \cite{kokil2020despeckling}, UNet \cite{mohamed2023ultrasound}, GAN \cite{zhang2022ultrasound}, autoencoder \cite{karaouglu2022removal, singh2023ultrasonic}, DDPM \cite{asgariandehkordi2024denoising, zhang2023ultrasound}, and customized attention mechanisms \cite{yancheng2023red}. Their training paradigms can be categorized as supervised and unsupervised learning.

\subsubsection{Supervised denoising} 
Supervised ultrasound denoising relies on large datasets of noisy-clean pairs to pretrain the model. However, since collecting \textit{in vivo} clean images is almost impractical, researchers have resorted to alternative label sources, including simulation \cite{hyun2019beamforming, yu2023deep, jiang2023controllable}, conventional algorithms \cite{cammarasana2022real}, images compounded from more tilted angles \cite{zhang2023ultrasound, asgariandehkordi2024denoising}, and other modalities \cite{zhang2023feature, wu2015echocardiogram, jarosik2021pixel}. However, none of these surrogates fully replicate the complex and dynamic \textit{in vivo} conditions, thus inevitably suffering from domain shift.

\subsubsection{Self-supervised denoising} 

Recently, the Noise2Noise (N2N) paradigm \cite{lehtinen2018noise2noise} bypasses clean ground truth by training on paired noisy observations with independent, zero-mean noise. In ultrasound, existing works constructed noisy pair datasets by exploiting physical data redundancies in multi-angle plane-wave (PW) ultrasound or algorithmic perturbations. Huang et al. \cite{huang2025self} divided PWs into odd and even groups as noisy pairs. Jung et al. \cite{jung2024unsupervised} used PWs from different angles as paired noisy speckle observations. Li et al. \cite{li2025speckle2self} proposed a multi-scale perturbation to create noisy pairs with variable speckle patterns. Yu et al. \cite{yu2025self} generated noisy pairs by sampling B-mode images from overspreading chunks. In addition to the N2N paradigm, Huh et al. \cite{huh2023tunable} used unmatched 2D references to guide tunable 3D image enhancement; Muth et al. \cite{muth2011unsupervised} achieved color-Doppler denoising via recursive dealiasing after segmentation. 

However, those methods assumed identical training and testing distributions, rendering them vulnerable to domain shifts across varying \textit{in vivo} environments. Besides, both supervised and self-supervised methods require large datasets and costly pretraining. Scaling up data collection and pretraining for every potential imaging domain is clinically infeasible. 

To achieve single-image denoising, DIP (Deep image prior) \cite{ulyanov2018deep} proved that a CNN could serve as an implicit denoiser. N2S (Noise2Self) \cite{batson2019noise2self} reduced independent noise across different measurements of one image. Nonetheless, prior studies mostly focused on either predicting clean structures or modeling noise representations. By jointly learning disentangled anatomical and noise representations, we overcome this limitation for more reliable denoising.

\begin{figure*}
\centering
\includegraphics[width=\textwidth]{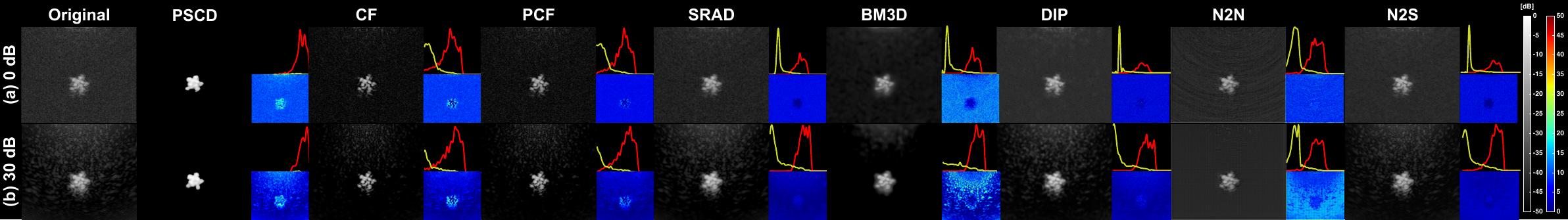}
\caption{Denoising results in simulations under (a) 0 dB and (b) 30 dB SNR. For each method, the denoised image is shown on the left; the upper-right shows its intensity probability density functions (PDFs) of the signal (red) and noise (yellow) regions; the lower-right shows its difference map compared with the original image. All PDFs are displayed in a lateral axis covering the dynamic range and a vertical axis from 0 to 1.}
\label{simu_compare}
\end{figure*}

\section{Methodology}
\subsection{Problem Statement}

As defined in Fig. \ref{solution}(a), in SAU imaging, an array transducer is divided into $N\geq2$ sub-apertures. Given $N$ times sequential sub-aperture transmissions and full-aperture reception, the received channel data are independently beamformed into $X=[x_1,\ldots,x_N]^{\mathsf T}\in\mathbb{C}^{N\times H\times W}$, where $H$ and $W$ denote the axial and lateral image dimensions, respectively. $x_n\in\mathbb{C}^{H\times W}$ is the original complex IQ frame measured by the $n$th sub-aperture event. $N$ sub-aperture transmissions provide distinct noisy measurements of the same tissue scattering distribution but exhibit aperture-dependent degradation mainly because of different virtual-source locations. Assume the ideal IQ signal representing the underlying tissue echogenicity is $y\in\mathbb{C}^{H\times W}$, the phenomenological degradation model can be formulated as
\begin{equation}
\label{eq:noise}
    x_n
    =
    y\odot\eta_n^{sp}
    +\eta_n^{sl}
    +\eta_n^{e},
    \qquad n=1,\ldots,N,
\end{equation}
where $\eta_n^{sp}$ denotes the tissue-dependent speckle modulation, $\eta_n^{sl}$ denotes the side-lobe component associated with the $n$th sub-aperture location, $\eta_n^{e}\sim\mathcal{N_C}(0,\sigma^2)$ denotes the electronic noise following complex normal distribution, and $\odot$ denotes the Hadamard product. 

Our objective is to estimate the clean IQ data $y$ from the measured IQ tensor $X$. Although this inverse problem is intrinsically ill-posed, the shared anatomical tissue characteristics across multiple sub-aperture observations provide the statistical redundancy to constrain the approximation of $y$. The final B-mode image is obtained by the standard envelope detection and logarithmic compression process on the estimated IQ signal $y$:
\begin{equation}
\label{eq:sau_bmode}
I = \operatorname{clip}_{[-D,0]}
\left[
20\log_{10}
\left(
\frac{
\left|y\right|
}{\max
\left|y\right|
}
\right)
\right],
\end{equation}
where $\operatorname{clip}(\cdot)$ limits its argument to $D$ dB dynamic range, and $I\in[-D,0]^{H\times W}$ is the displaced B-mode image.

\subsection{Cross-aperture Signal Coherence and Incoherence}

The $N$ independently beamformed IQ frames provide multiple noisy measurements of the same tissue region. Here we generalize the cross-aperture coherence theory applied at the pre-beamformed RF signals into the beamformed IQ domain to reduce data and computational complexity. Fig. \ref{solution}(a, bottom) shows an example of the coherence and incoherence maps using this generalized definition.

Let the local average of a 2D window $\Omega(\bm p)$ centered at a pixel $\bm p$ denoted by
\begin{equation}
\label{average}
    \langle x_n\rangle_{\bm p}
    =\sum_{\bm q\in\Omega(\bm p)}w_{\bm p,\bm q}x_n(\bm q),
    \quad
    \sum_{\bm q\in\Omega(\bm p)}w_{\bm p,\bm q}=1,
\end{equation}
where $w_{\bm p,\bm q}\geq0$ is the normalized weight. For a pair $(n,m)$ of sub-aperture IQ frames with nonzero local energy, their normalized local correlation is
\begin{equation}
\label{coherence}
    \gamma_{nm}(\bm p)=
    \frac{\langle x_n x_m^*\rangle_{\bm p}}
    {\sqrt{\langle|x_n|^2\rangle_{\bm p}
                   \langle|x_m|^2\rangle_{\bm p}}},
\end{equation}
where $(\cdot)^*$ denotes complex conjugation \cite{Hyun_Crowley_Dahl_2017, Long_Trahey_Bottenus_2022}. 

A spatial coherence map is defined as the squared magnitude of Eq. \eqref{coherence} over all $\bm p$ pixels throughout the IQ frame: $\mathcal{C}_{nm}(\bm p) = |\gamma_{nm}(\bm p)|^2 \in [0, 1]$. It quantifies the local similarity between complex IQ patterns, accounting for both amplitude and phase. The complementary pairwise incoherence is thus defined as
\begin{equation}
    \mathcal{I}_{nm}(\bm p)
    =
    1-\mathcal{C}_{nm}(\bm p), \qquad 0\leq\mathcal{I}_{nm}\leq1.
    \label{incoherence}
\end{equation}

High cross-aperture coherence indicates that a local IQ pattern is shared under different sub-aperture transmissions. As shown in Fig. \ref{solution}(a, bottom), high coherence values arise from underlying tissue backscatter in overlapping transmit fields and sufficiently consistent focusing. Conversely, the incoherent part indicates aperture-dependent variation caused by side lobes, off-axis scattering, or phase aberration. However, $\mathcal{C}_{nm}$ characterizes only a second-order relationship between two local IQ patches. Such a prescribed spatial window and weighting rule cannot sufficiently characterize the multiscale spatial coupling among tissue echoes and composite aperture-dependent degradation. This motivates extending the explicit IQ-domain coherence statistic to learned multiscale representations using complex-valued convolutional neural networks.

\begin{table}[]
\centering
\caption{Parameters used in \textit{in vivo} imaging and k-Wave simulation.}
\scalebox{1}{
\begin{tabular}{@{}ccc@{}}
\toprule
\multicolumn{2}{c}{\textbf{Imaging parameter}}            & \textbf{Value}           \\ \midrule
\multicolumn{2}{c}{Probe}                                 & P4-2 phased array                    \\
\multicolumn{2}{c}{Element number}                                 & 64                    \\
\multicolumn{2}{c}{Virtual source}                        & 1.28 mm behind the array \\
\multicolumn{2}{c}{Transmit wave}                            & Diverging wave         \\
\multicolumn{2}{c}{Opening angle}                            & ${\pi}/{2}$         \\
\multicolumn{2}{c}{Transmit frequency}                    & 2.5 MHz                  \\ \midrule
\multirow{3}{*}{\textit{\textbf{In vivo}}} & Sub-aperture number & N=2                                                    \\
                                           & Excitation      & CaSA \cite{zhang2017ultrafast} driven at 4V \\
                                     & Frame rate          & 1600 fps                 \\ \midrule
\multirow{2}{*}{\textbf{Simulation}} & Sub-aperture number    & N=4, 8                     \\
                                     & SNR                & 0, 10, 20, 30 dB         \\ \bottomrule
\end{tabular}}
\label{parameter}
\end{table}

\begin{table}[]
\centering
\caption{Medium properties used in k-Wave simulation.}
\scalebox{0.8}{
\setlength{\tabcolsep}{14pt}
\begin{tabular}{@{}ccc@{}}
\toprule
\multirow{2}{*}{\makecell{\textbf{Medium} \\ \textbf{parameter}}} & \textbf{Background} & \textbf{Inclusion}      \\ \cmidrule(l){2-3} 
                                 & \textbf{Water}      & \textbf{Cardiac muscle \cite{mast2000empirical}} \\ \midrule
Speed of sound: $c_0$ (m/s)                             & 1540                & 1545                    \\
Density: $\rho \ (kg/m^3)$                             & 1000                & 1060                    \\
Nonlinearity: B/A                           & 8                   & 7.1                     \\
Absorption: $\alpha_0$ (dB/MHz/cm)                             & 0.3                 & 0.52                    \\
Geometry                      & /                   & Circle, star            \\ \bottomrule
\end{tabular}}
\label{medium}
\end{table}

\begin{figure*}
\centering
\includegraphics[width=\textwidth]{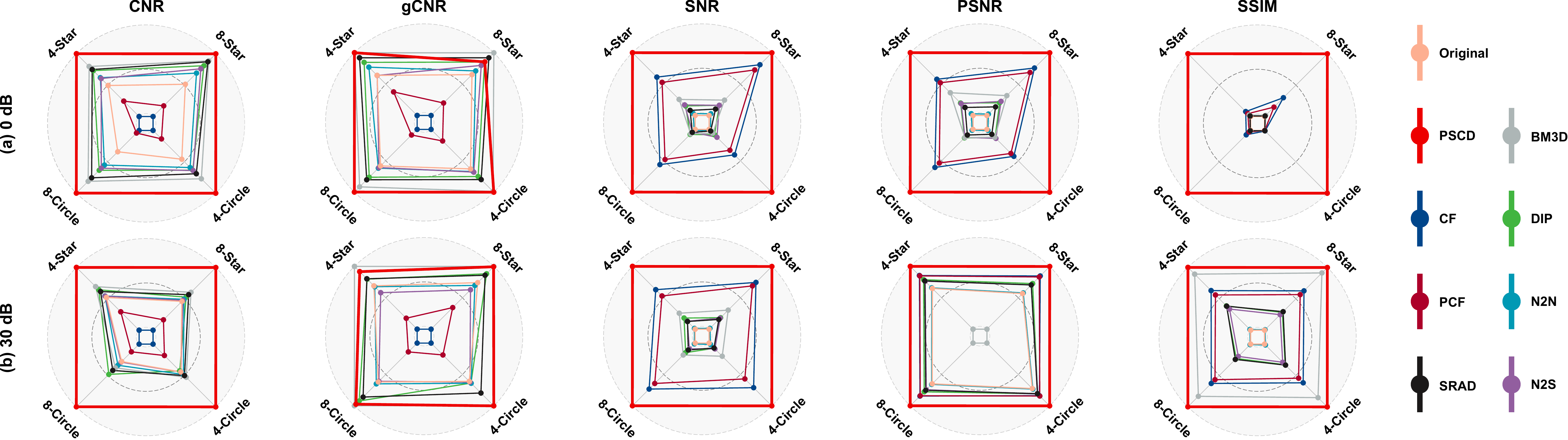}
\caption{The radar plots of CNR, gCNR, SNR, PSNR, and SSIM on simulation data under two noise levels. For each metric, results for N=4 or N=8 apertures and star or circle inclusion are shown at the four corners, respectively.}
\label{simu_radar}
\end{figure*}

\subsection{Pyramid Self-contrastive Disentanglement}

Rather than directly treating coherence and incoherence maps as clean and noise masks, we use them as physical priors for implicit disentanglement and leverage data-adaptive neural networks to learn the cross-aperture coherence and incoherence components. After multiple transmit-receive events using synthetic transmit apertures, our PSCD processes the received noisy IQ tensor in two stages: (1) adaptive coherence disentanglement and (2) selective clean IQ reconstruction. The overall framework is illustrated in Fig. \ref{model}(a), where our model consists of an anatomy encoder ($f_a$), a noise encoder ($f_n$), and a decoder ($f_d$). For single-shot SAU imaging, the originally measured $X$ is encoded into two complementary latent representations, termed the anatomy space and the noise space. The anatomy space is constrained to retain features shared across sub-aperture measurements, whereas the noise space is for aperture-dependent degradations.

We design two operational modes for the framework:
\begin{align}
\mathrm {Training:} \ \ \widetilde{X} & = f_d(f_a(X)+f_n(X)), \label{tilde} \\
\mathrm {Inference:} \ \ Y &= f_d(f_a(X)). \label{hat}
\end{align}
As in Eq. \ref{tilde}, $f_d$ reconstructs another randomly sampled noisy variant $\widetilde{X}\in\mathbb C^{N\times H\times W}$ by fusing the two spaces. This noisy reconstruction serves as a proxy task for Test-Time Training (TTT, Section E). Once optimized, as shown in the orange block in Fig. \ref{model}(a) and Eq. \ref{hat}, only the coherence-bearing anatomy space is decoded to reconstruct a clean estimation $Y\in\mathbb C^{N\times H\times W}$ during the inference phase, while the incoherent space is excluded from this coherent decoding process. Before B-mode image formation, we coherently compound $Y$ along the sub-aperture dimension to obtain the desired IQ image $y$.

\subsubsection{Random noisy sampling}

The PSCD disentanglement is illustrated in Fig. \ref{model}(c). In each iteration, PSCD randomly permutes the original tensor $X$ along the aperture dimension to generate $M$ noisy views with different
sub-aperture orders. Let $\mathfrak S_N$ denote the set of all $N!$ permutations of $\{1,\ldots,N\}$. 

PSCD first samples a base permutation $\pi\in\mathfrak{S}_N$ at random. For a cyclic-shift index $i\in\{0,\ldots,N-1\}$, define
\begin{equation}
    \pi_{i}(r) =
    \pi\!\left(
        1+(r+i-1)\bmod N
    \right),
    \label{eq:cyclic_permutation}
\end{equation}
where $r\in\{1,\ldots,N\}$ is the channel position and $\bmod$ returns the remainder. PSCD then generates $\pi_{1},...,\pi_{M}$ using distinct shift indices and construct the multi-view set as
\begin{equation}
    \mathcal{X}:=\{X_i=P_{\pi_i}X \ | \ i=1,\ldots,M\},
    \qquad 2\leq M\leq N,
    \label{eq:views}
\end{equation}
where $P_{\pi_i}\in\{0,1\}^{N\times N}$ is the permutation matrix of $\pi_i$. For any view pair $(X_i,X_j)$ with $i\neq j$, any channel at the same position originates from different sub-apertures.

\subsubsection{Coherence and incoherence encoding}

As shown in Fig. \ref{model}(c), $f_a$ and $f_n$ consist of $K$ layers with progressively enlarged receptive fields. Using $X_i$ as input, $f_a$ and $f_n$ encode two feature pyramids:
\begin{align}
\mathcal{A}_{i} := & \left \{ \mathcal{A}_{i}^k \mid k=1 ,\dots, K\right \} =f_{\mathrm {N}}(f_a(X_{i})), \\
\eta_{i} := & \left \{ \eta_{i}^k \mid k=1 ,\dots, K\right \}=f_{\mathrm {N}}(f_n(X_{i})), \\
f_{\mathrm {N}}(\cdot)= &\frac{\operatorname{vec}(\cdot)}{\|\operatorname{vec}(\cdot)\|_2}:\mathbb{C}^{N_k\times H_k \times W_k} \mapsto \mathbb{C}^{N_k H_k W_k},
\end{align}
where the $L_2$ normalization $f_{\mathrm {N}}$ flattens 2D feature tensors into 1D dense embeddings and aligns them into a unit hypersphere; $\mathcal{A}^k$ and $\eta^k$ denote dense embeddings from the $k$-th layer in $f_a$ and $f_n$, respectively.

Instead of using a predefined local average as in Eq. \ref{average} to estimate coherence, PSCD learns the local weights using complex convolutional neural networks. Let $X=X_{\mathrm{real}}+jX_{\mathrm{imag}}$ be the input of a convolution operation. The output $X'$ is calculated as
\begin{align}
\begin{bmatrix} 
X'_{\mathrm{real}} \\ 
X'_{\mathrm{imag}} 
\end{bmatrix} 
= f_{NL}(\begin{bmatrix} 
\mathbf{W}_{\mathrm{real}} \ast X_{\mathrm{real}} - \mathbf{W}_{\mathrm{imag}} \ast X_{\mathrm{imag}} \\ 
\mathbf{W}_{\mathrm{real}} \ast X_{\mathrm{imag}} + \mathbf{W}_{\mathrm{imag}} \ast X_{\mathrm{real}}
\end{bmatrix}),
\label{conv}
\end{align}
where $f_{NL}$ is the nonlinear function including a complex batch normalization and a complex PReLU activation; $\mathbf{W}_{\mathrm{real}}$ and $\mathbf{W}_{\mathrm{imag}}$ are parameter sets for the real and imaginary parts of the network, respectively. Each resulting $X' \subset \mathcal{A}_i \ \mathrm {or} \ \eta_i$ can be regarded as a nonlinear representation of the input $X$.

After every convolutional layer, a magnitude-based complex max-pooling downsamples the feature maps to $1/2$ size. It progressively enlarges the receptive fields throughout the encoder and extends the fixed 2D local window $\Omega(\bm p)$ in Eq. \ref{average} into a multi-scale pyramid. In this way, PSCD encodes all $M$ noisy views and extract coherence space $\{\mathcal{A}_i | \ i=1, \dots, M\}$ and incoherence space $\{\eta_i | \ i=1, \dots, M\}$ in a data-adaptive manner.

\begin{figure*}
\centering
\includegraphics[width=\textwidth]{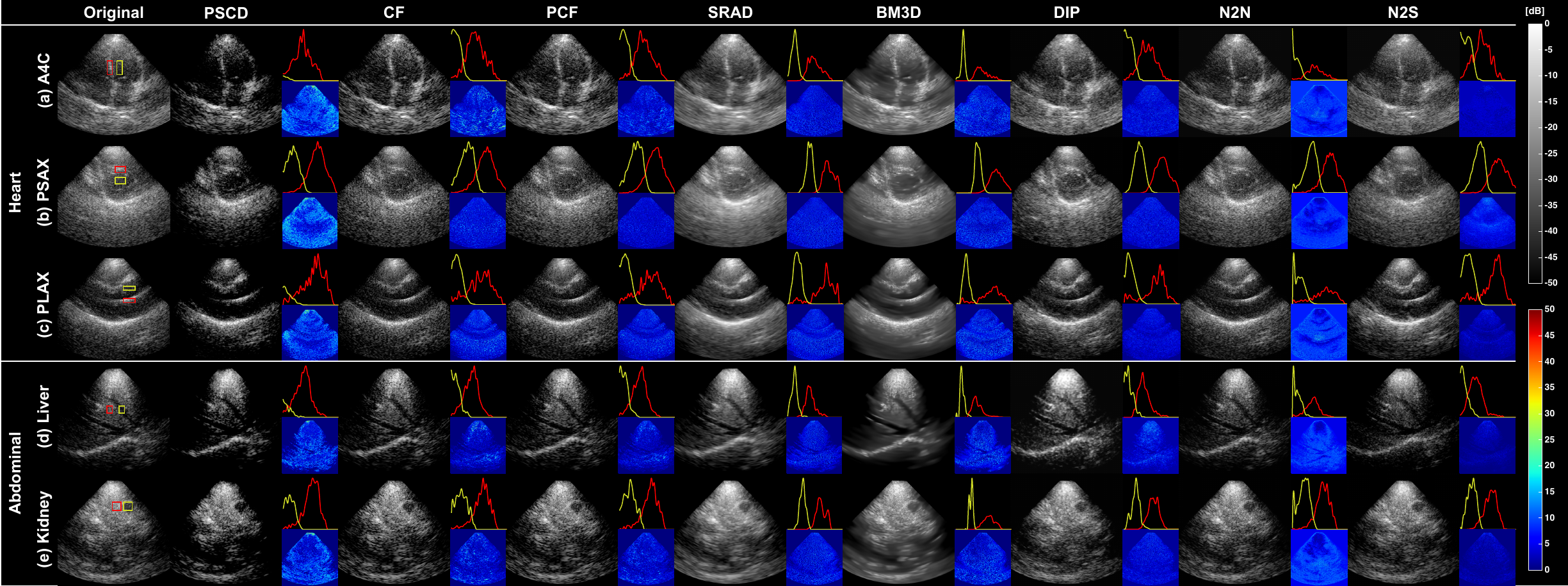}
\caption{Denoising comparison on \textit{in vivo} human heart in (a) A4C, (b) PSAX, and (c) PLAX views, and \textit{in vivo} abdominal organs of (d) liver and (e) kidney. The red and yellow boxes denote the signal and noise regions to calculate metrics. The visualization layout is identical to that in Fig. \ref{simu_compare}.}
\label{invivo_compare}
\end{figure*}

\subsubsection{Coherence-guided disentanglement}

The coherence operator $\mathcal{C}_{nm}$ can be reformulated as a form of complex cosine similarity. By vectorizing the locally weighted IQ samples within \(\Omega(\bm p)\) as \(\bm x_n =\left[\sqrt{w_{p,q}} \ x_n(q)\right]_{q\in\Omega(p)}\),
\begin{equation}
\mathcal{C}_{nm}(\bm p) = \mathrm {Cos} (\bm x_m,\bm x_n) = \frac{|\bm x_m^{\mathrm H} \cdot \bm x_n|^2} {\|\bm x_n\|_F^2 \cdot \|\bm x_m\|_F^2},
 \label{complex_cosine}
\end{equation}
where $(\cdot)^{\mathsf H}$ denotes Hermitian transpose and $\|\cdot\|_F$ denotes Frobenius norm. Similarly, we generalize Eq. \ref{complex_cosine} as a coherence-based critic on the feature domain instead of on the IQ signal domain. The PSCD disentanglement can thus be formulated as a dense contrastive learning process \cite{wang2021dense}.

At each pyramid layer, for any $i\neq j$, the first objective attracts $\mathcal{A}_i^k$ and $\mathcal{A}_j^k$ to optimize $f_a$, extracting coherence features between any sub-aperture pair $(X_i, X_j)$. Let the set of all unordered pairs among the $M$ distinct views as $Q_M:=\{(i,j):1\leq i<j\leq M\}$. The first objective can be formulated as
\begin{equation}
\begin{aligned}
\max_{f_a} \ \sum_{k=1}^K\sum_{Q_M} \mathrm {Cos} (\mathcal{A}_i^k, \mathcal{A}_j^k) &
= \sum_{k=1}^K\sum_{Q_M} \frac{|(\mathcal{A}_i^k)^{\mathrm H} \cdot \mathcal{A}_j^k|^2}{\left \| \mathcal{A}_i^k \right \|_F^2 \cdot \left \| \mathcal{A}_j^k \right \|_F^2}, \label{max_a12}
\end{aligned}
\end{equation}
where $\mathrm {Cos} (\mathcal{A}_1^k, \mathcal{A}_2^k)$ can be interpreted as the feature-domain coherence using the $k$th pyramid local window.

According to Eq. \ref{incoherence}, maximizing $\mathcal{I}_{nm}$ equals to minimizing $\mathcal{C}_{nm}$. Thus, by minimizing the complex cosine similarity, PSCD's second objective is to repel $\eta_1^k$ and $\eta_2^k$ to optimize $f_n$ to extract the incoherence features:
\begin{align}
\min_{f_n} \ \sum_{k=1}^K\sum_{Q_M} \mathrm {Cos} (\eta_i^k, \eta_j^k) =
\sum_{k=1}^K\sum_{Q_M} \frac{|(\eta_i^k)^{\mathrm H} \cdot \eta_j^k|^2}{\left \| \eta_i^k \right \|_F^2 \cdot \left \| \eta_j^k \right \|_F^2}. \label{min_n12}
\end{align}

At each pyramid layer, for every random sampled $X_i$, PSCD also repels coherence features $\mathcal{A}_{i}$ from its intra-sample incoherence features $\eta_{i}$ to orthogonalize two feature spaces, reducing weight redundancy between $f_a$ and $f_n$: 
\begin{align}
\min_{f_a, f_n} \ \sum_{k=1}^K \sum_{i=1}^M \mathrm {Cos} (\mathcal{A}_{i}^k, \eta_{i}^k) = 
\sum_{k=1}^K \sum_{i=1}^M \frac{|(\mathcal{A}_{i}^k)^{\mathrm H} \cdot \eta_{i}^k|^2}{\left \| \mathcal{A}_{i}^k \right \|_F^2 \cdot \left \| \eta_{i}^k \right \|_F^2}. \label{min_an}
\end{align}

Inspired by InfoNCE \cite{oord2018representation} and given $\|\mathcal A_i^k\|_F=\|\eta_i^k\|_F=1$, we rewrite the multiple objectives of this dense contrastive learning as an aggregated cross-entropy loss: 
\begin{equation}
\begin{aligned}
& \mathcal{L}_{\mathrm{con}}= 
-\textstyle \sum_{k=1}^{K} \frac{1}{N_k} \log\\
& \frac{\sum_{Q_M}e^{\frac{|(\mathcal{A}_i^k)^{\mathrm H}   \mathcal{A}_j^k|^2}{\tau}}}
{\sum_{Q_M} \left (e^{\frac{|(\mathcal{A}_i^k)^{\mathrm H}   \mathcal{A}_j^k|^2}{\tau}} + 
e^{\frac{|(\eta_i^k)^{\mathrm H}   \eta_j^k|^2}{\tau}} \right ) + 
\sum_{i=1}^M
e^{\frac{|(\mathcal{A}_{i}^k)^{\mathrm H}   \eta_{i}^k|^2}{\tau}}},
\end{aligned}
\label{lcon}
\end{equation}
where $|Q_M|=\binom{M}{2}=\frac{M(M-1)}{2}$ and $\tau$ is the temperature parameter. All $(\mathcal{A}_i^k, \mathcal{A}_j^k)$ construct $K |Q_M|$ positive pairs and all $(\eta_i^k, \eta_j^k)$ and $(\mathcal{A}_i^k, \eta_i^k)$ construct $K (|Q_M| + M)$ negative pairs. Different from conventional contrastive learning methods, $\mathcal{L}_{con}$ learns from multiple noisy views of single-shot data $X$ without relying on external negative samples.

We define a $3 \times 3$ kernel size and a $2\times$ max pooling after each layer to construct $f_a$ and $f_n$ using $K=3$. The convolutional channel numbers are $N_1, N_2, N_3=16, 32, 64$. This pyramid structure extends a fixed local coherence map as in Eq. \ref{coherence} into multiscale coherence and incoherence spaces, thereby facilitating the disentanglement of diverse noise patterns in different scales while ensuring cross-scale anatomical consistency.

\subsection{Aperture-to-Aperture Test-time Training}

\begin{table*}[]
\caption{Denoising performance of PSCD against 7 comparison methods on simulation data under 4 electronic noise levels.}
\label{simu_metric}
\centering
\scalebox{0.85}{
\setlength{\tabcolsep}{7pt}
\begin{tabular}{@{}cc|ccccccccc@{}}
\toprule
 &
   &
   &
  \multicolumn{8}{c}{\textbf{Denoising Methods {\scriptsize($\uparrow$ or $\downarrow$ in Percentage)}}} \\ \cmidrule(l){4-11} 
\multirow{-2}{*}{\textbf{Metrics}} &
  \multirow{-2}{*}{\textbf{Noise Level}} &
  \multirow{-2}{*}{\makecell{\textbf{Original} \\ \textbf{Image}}} &
  \textbf{PSCD} &
  \textbf{CF} &
  \textbf{PCF} &
  \textbf{SRAD} &
  \textbf{BM3D} &
  \textbf{DIP} &
  \textbf{N2N} &
  \textbf{N2S} \\ \midrule
 &
  \textit{0 dB} &
  6.56 &
  9.29 {\scriptsize (42$\%\uparrow$)} &
  4.23 {\scriptsize (35$\%\downarrow$)} &
  5.04 {\scriptsize (23$\%\downarrow$)} &
  8.06 {\scriptsize (23$\%\uparrow$)} &
  8.33 {\scriptsize (27$\%\uparrow$)} &
  7.77 {\scriptsize (19$\%\uparrow$)} &
  7.34 {\scriptsize (12$\%\uparrow$)} &
  7.51 {\scriptsize (14$\%\uparrow$)} \\
 &
  \textit{10 dB} &
  7.01 &
  8.42 {\scriptsize (20$\%\uparrow$)} &
  4.85 {\scriptsize (31$\%\downarrow$)} &
  5.78 {\scriptsize (18$\%\downarrow$)} &
  7.78 {\scriptsize (11$\%\uparrow$)} &
  7.71 {\scriptsize (10$\%\uparrow$)} &
  7.75 {\scriptsize (11$\%\uparrow$)} &
  7.40 {\scriptsize (5$\%\uparrow$)} &
  7.19 {\scriptsize (3$\%\uparrow$)} \\
 &
  \textit{20 dB} &
  7.00 &
  9.58 {\scriptsize (37$\%\uparrow$)}&
  4.93 {\scriptsize (30$\%\downarrow$)}&
  5.87 {\scriptsize (16$\%\downarrow$)}&
  7.48 {\scriptsize (7$\%\uparrow$)}&
  7.60 {\scriptsize (9$\%\uparrow$)}&
  7.36 {\scriptsize (5$\%\uparrow$)}&
  7.21 {\scriptsize (3$\%\uparrow$)}&
  7.05 {\scriptsize (1$\%\uparrow$)}\\
\multirow{-4}{*}{\textit{\textbf{CNR}}} &
  \textit{30 dB} &
  6.99 &
  9.62 {\scriptsize (37$\%\uparrow$)} &
  4.93 {\scriptsize (30$\%\downarrow$)} &
  5.86 {\scriptsize (16$\%\downarrow$)} &
  7.45 {\scriptsize (6$\%\uparrow$)} &
  7.60 {\scriptsize (9$\%\uparrow$)} &
  7.39 {\scriptsize (6$\%\uparrow$)} &
  7.17 {\scriptsize (3$\%\uparrow$)} &
  7.05 {\scriptsize (1$\%\uparrow$)} \\ \midrule 
 &
  \textit{0 dB} &
  0.845 &
  0.896 {\scriptsize (6$\%\uparrow$)} &
  0.742 {\scriptsize (12$\%\downarrow$)} &
  0.776 {\scriptsize (8$\%\downarrow$)} &
  0.877 {\scriptsize (4$\%\uparrow$)} &
  0.897 {\scriptsize (6$\%\uparrow$)} &
  0.868 {\scriptsize (3$\%\uparrow$)} &
  0.852 {\scriptsize (1$\%\uparrow$)} &
  0.849 {\scriptsize (-)} \\
 &
  \textit{10 dB} &
  0.848 &
  0.866 {\scriptsize (2$\%\uparrow$)} &
  0.761 {\scriptsize (10$\%\downarrow$)} &
  0.791 {\scriptsize (7$\%\downarrow$)} &
  0.877 {\scriptsize (3$\%\uparrow$)} &
  0.896 {\scriptsize (6$\%\uparrow$)} &
  0.873 {\scriptsize (3$\%\uparrow$)} &
  0.852 {\scriptsize (-)} &
  0.844 {\scriptsize (1$\%\downarrow$)} \\
 &
  \textit{20 dB} &
  0.848 &
  0.899 {\scriptsize (6$\%\uparrow$)} &
  0.762 {\scriptsize (10$\%\downarrow$)} &
  0.792 {\scriptsize (7$\%\downarrow$)} &
  0.876 {\scriptsize (3$\%\uparrow$)} &
  0.896 {\scriptsize (6$\%\uparrow$)} &
  0.862 {\scriptsize (2$\%\uparrow$)} &
  0.853 {\scriptsize (1$\%\uparrow$)} &
  0.841 {\scriptsize (1$\%\downarrow$)} \\
\multirow{-4}{*}{\textit{\textbf{gCNR}}} &
  \textit{30 dB} &
  0.848 &
  0.895 {\scriptsize (6$\%\uparrow$)} &
  0.762 {\scriptsize (10$\%\downarrow$)} &
  0.792 {\scriptsize (7$\%\downarrow$)} &
  0.876 {\scriptsize (3$\%\uparrow$)} &
  0.895 {\scriptsize (6$\%\uparrow$)} &
  0.869 {\scriptsize (2$\%\uparrow$)} &
  0.852 {\scriptsize (-)} &
  0.844 {\scriptsize (-)} \\ \midrule 
 &
  \textit{0 dB} &
  5.83 &
  12.36 {\scriptsize (112$\%\uparrow$)} &
  9.69 {\scriptsize (66$\%\uparrow$)} &
  9.17 {\scriptsize (57$\%\uparrow$)} &
  6.24 {\scriptsize (7$\%\uparrow$)} &
  6.83 {\scriptsize (17$\%\uparrow$)} &
  6.60 {\scriptsize (13$\%\uparrow$)} &
  5.98 {\scriptsize (3$\%\uparrow$)} &
  6.63 {\scriptsize (14$\%\uparrow$)} \\
 &
  \textit{10 dB} &
  7.25 &
  11.81 {\scriptsize (63$\%\uparrow$)} &
  10.51 {\scriptsize (45$\%\uparrow$)} &
  10.11 {\scriptsize (39$\%\uparrow$)} &
  7.83 {\scriptsize (8$\%\uparrow$)} &
  8.51 {\scriptsize (17$\%\uparrow$)} &
  8.10 {\scriptsize (12$\%\uparrow$)} &
  7.38 {\scriptsize (2$\%\uparrow$)} &
  7.99 {\scriptsize (10$\%\uparrow$)} \\
 &
  \textit{20 dB} &
  7.63 &
  11.29 {\scriptsize (48$\%\uparrow$)} &
  10.60 {\scriptsize (39$\%\uparrow$)} &
  10.21 {\scriptsize (34$\%\uparrow$)} &
  8.14 {\scriptsize (7$\%\uparrow$)} &
  8.67 {\scriptsize (14$\%\uparrow$)} &
  8.26 {\scriptsize (8$\%\uparrow$)} &
  7.70 {\scriptsize (1$\%\uparrow$)} &
  8.16 {\scriptsize (7$\%\uparrow$)} \\
\multirow{-4}{*}{\textit{\textbf{SNR}}} &
  \textit{30 dB} &
  7.67 &
  11.84 {\scriptsize (54$\%\uparrow$)} &
  10.60 {\scriptsize (38$\%\uparrow$)} &
  10.21 {\scriptsize (33$\%\uparrow$)} &
  8.17 {\scriptsize (6$\%\uparrow$)} &
  8.68 {\scriptsize (13$\%\uparrow$)} &
  8.31 {\scriptsize (8$\%\uparrow$)} &
  7.73 {\scriptsize (1$\%\uparrow$)} &
  8.19 {\scriptsize (7$\%\uparrow$)}  \\ \midrule 
 &
  \textit{0 dB} &
  12.48 &
  24.10 {\scriptsize (93$\%\uparrow$)} &
  19.37 {\scriptsize (55$\%\uparrow$)} &
  18.66 {\scriptsize (50$\%\uparrow$)} &
  13.73 {\scriptsize (10$\%\uparrow$)} &
  15.16 {\scriptsize (21$\%\uparrow$)} &
  14.40 {\scriptsize (15$\%\uparrow$)} &
  12.71 {\scriptsize (2$\%\uparrow$)} &
  14.46 {\scriptsize (16$\%\uparrow$)} \\
 &
  \textit{10 dB} &
  16.15 &
  22.77 {\scriptsize (41$\%\uparrow$)} &
  21.29 {\scriptsize (32$\%\uparrow$)} &
  21.14 {\scriptsize (31$\%\uparrow$)} &
  18.21 {\scriptsize (13$\%\uparrow$)} &
  21.43 {\scriptsize (33$\%\uparrow$)} &
  18.90 {\scriptsize (17$\%\uparrow$)} &
  16.51 {\scriptsize (2$\%\uparrow$)} &
  18.65 {\scriptsize (15$\%\uparrow$)} \\
 &
  \textit{20 dB} &
  17.91 &
  23.33 {\scriptsize (30$\%\uparrow$)} &
  21.47 {\scriptsize (20$\%\uparrow$)} &
  21.42 {\scriptsize (20$\%\uparrow$)} &
  19.86 {\scriptsize (11$\%\uparrow$)} &
  21.89 {\scriptsize (22$\%\uparrow$)} &
  20.16 {\scriptsize (13$\%\uparrow$)} &
  18.13 {\scriptsize (1$\%\uparrow$)} &
  19.70 {\scriptsize (10$\%\uparrow$)} \\
\multirow{-4}{*}{\textit{\textbf{PSNR}}} &
  \textit{30 dB} &
  18.19 &
  23.95 {\scriptsize (32$\%\uparrow$)} &
  21.44 {\scriptsize (18$\%\uparrow$)} &
  21.36 {\scriptsize (17$\%\uparrow$)} &
  20.00 {\scriptsize (10$\%\uparrow$)} &
  22.05 {\scriptsize (21$\%\uparrow$)} &
  20.28 {\scriptsize (11$\%\uparrow$)} &
  18.34 {\scriptsize (1$\%\uparrow$)} &
  19.97 {\scriptsize (10$\%\uparrow$)} \\ \midrule 
 &
  \textit{0 dB} &
  0.195 &
  0.982 {\scriptsize (403$\%\uparrow$)} &
  0.286 {\scriptsize (46$\%\uparrow$)} &
  0.239 {\scriptsize (22$\%\uparrow$)} &
  0.201 {\scriptsize (3$\%\uparrow$)} &
  0.207 {\scriptsize (6$\%\uparrow$)} &
  0.201 {\scriptsize (3$\%\uparrow$)} &
  0.197 {\scriptsize (1$\%\uparrow$)} &
  0.201 {\scriptsize (3$\%\uparrow$)} \\
 &
  \textit{10 dB} &
  0.204 &
  0.897 {\scriptsize (339$\%\uparrow$)} &
  0.597 {\scriptsize (192$\%\uparrow$)} &
  0.516 {\scriptsize (152$\%\uparrow$)} &
  0.240 {\scriptsize (17$\%\uparrow$)} &
  0.709 {\scriptsize (247$\%\uparrow$)} &
  0.244 {\scriptsize (19$\%\uparrow$)} &
  0.209 {\scriptsize (2$\%\uparrow$)} &
  0.248 {\scriptsize (21$\%\uparrow$)} \\
 &
  \textit{20 dB} &
  0.265 &
  0.980 {\scriptsize (269$\%\uparrow$)} &
  0.732 {\scriptsize (176$\%\uparrow$)} &
  0.688 {\scriptsize (159$\%\uparrow$)} &
  0.473 {\scriptsize (78$\%\uparrow$)} &
  0.876 {\scriptsize (230$\%\uparrow$)} &
  0.451 {\scriptsize (70$\%\uparrow$)} &
  0.275 {\scriptsize (4$\%\uparrow$)} &
  0.411 {\scriptsize (55$\%\uparrow$)} \\
\multirow{-4}{*}{\textit{\textbf{SSIM}}} &
  \textit{30 dB} &
  0.325 &
  0.979 {\scriptsize (201$\%\uparrow$)} &
  0.734 {\scriptsize (126$\%\uparrow$)} &
  0.691 {\scriptsize (113$\%\uparrow$)} &
  0.527 {\scriptsize (62$\%\uparrow$)} &
  0.895 {\scriptsize (176$\%\uparrow$)} &
  0.522 {\scriptsize (61$\%\uparrow$)} &
  0.331 {\scriptsize (2$\%\uparrow$)} &
  0.497 {\scriptsize (53$\%\uparrow$)} \\ \bottomrule
\end{tabular}
}
\end{table*}

\begin{figure*}
\centering
\includegraphics[width=\textwidth]{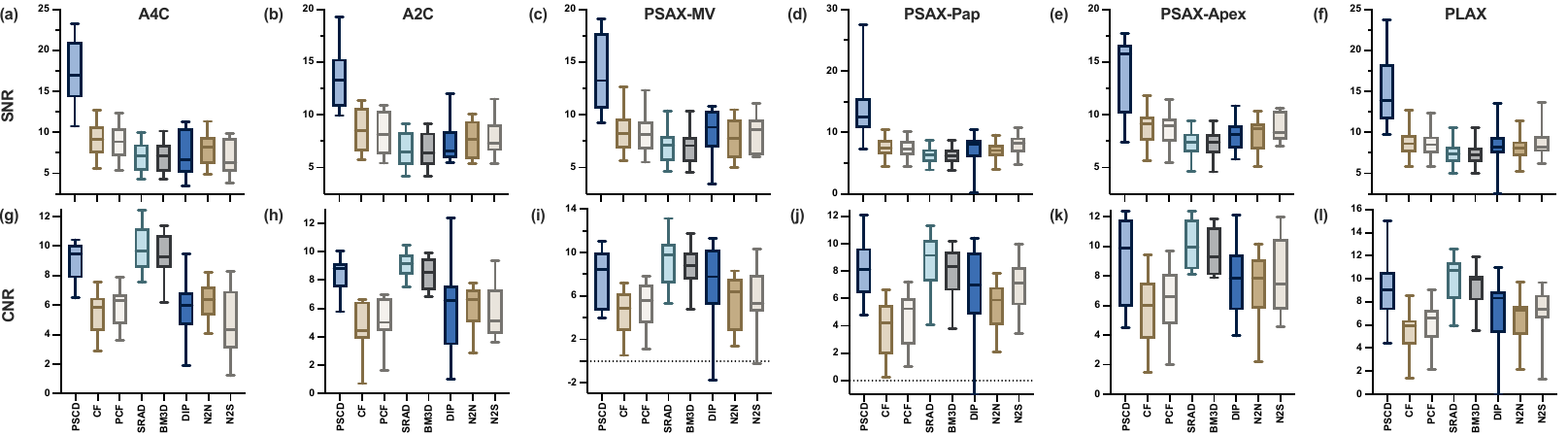}
\caption{The box plots of (a-f) SNR and (g-l) CNR calculated from the original image and the images obtained by PSCD and seven comparison methods on \textit{in vivo} echocardiograms in six different views (A4C, A2C, PSAX-MV, PSAX-Pap, PSAX-Apex, and PLAX).}
\label{echo_box}
\end{figure*}

Our framework regards each unpredictable testing environment as an independent data domain and conducts initialized TTT on it. This pretraining-free method eliminates the premise that training and testing data are drawn from the same domain, fundamentally solving the domain shift problem. As illustrated in Fig.~\ref{model}(b), we introduce an Aperture-to-Aperture (A2A) test-time training strategy to provide self-supervision for reconstruction fidelity. 

\subsubsection{Random noisy reconstruction}

Within each iteration, PSCD pairwise groups the multi-view set $\mathcal{X}$ to create (source view, target view) pairs. It first uniformly samples an offset $\delta\in \{1,\ldots,M-1\} $. The target view index for the $i$th source view is 
\begin{equation}
    j
    =1+\big((i+\delta-1)\bmod M\big),
    \qquad i\in\{1,\ldots,M\}.
    \label{eq:cyclic-target}
\end{equation}
In this way, every sampled view appears once as a source and once as a target without repetition. 

The decoder $f_d$ has an asymmetrical architecture compared to the encoder, except using transposed complex convolutions to upsample the feature maps. Skip connections between the encoder and decoder are implemented via complex summation. As in Eq. \ref{tilde}, given a random sample $X_i \in \mathcal{X}$, $f_d$ uses the fused representation by summing the anatomy space and the noise space in every skip connection to perform directional noisy reconstruction:
\begin{equation}
    \widetilde{X}_{j\mid i}
    =f_d\!\left(f_a(X_i)+f_n(X_i)\right),
    \label{eq:directional-reconstruction}
\end{equation}
where $\widetilde X_{j\mid i}$ denotes the prediction of target view $X_j \in \mathcal{X}$ from the input source view.

\begin{table*}[]
\caption{Denoising performance of PSCD against 7 comparison methods on simulation data with 2 inclusion geometries.}
\label{simu_geo}
\centering
\scalebox{0.8}{
\setlength{\tabcolsep}{7pt}
\begin{tabular}{@{}cc|ccccccccc@{}}
\toprule
 &
   &
   &
  \multicolumn{8}{c}{\textbf{Denoising Methods {\scriptsize($\uparrow$ or $\downarrow$ in Percentage)}}} \\ \cmidrule(l){4-11} 
\multirow{-2}{*}{\makecell{\textbf{Inclusion} \\ \textbf{Geometrics}}} &
  \multirow{-2}{*}{\textbf{Metrics}} &
  \multirow{-2}{*}{\makecell{\textbf{Original} \\ \textbf{Image}}} &
  \textbf{PSCD} &
  \textbf{CF} &
  \textbf{PCF} &
  \textbf{SRAD} &
  \textbf{BM3D} &
  \textbf{DIP} &
  \textbf{N2N} &
  \textbf{N2S} \\ \midrule
 &
  \textit{\textbf{CNR}} &
  8.09 &
  10.62 {\scriptsize (31$\%\uparrow$)}&
  5.50 {\scriptsize (32$\%\downarrow$)}&
  6.70 {\scriptsize (17$\%\downarrow$)}&
  8.93 {\scriptsize (10$\%\uparrow$)}&
  9.14 {\scriptsize (13$\%\uparrow$)}&
  8.92 {\scriptsize (10$\%\uparrow$)}&
  8.45 {\scriptsize (4$\%\uparrow$)}&
  8.39 {\scriptsize (4$\%\uparrow$)}\\
 &
  \textit{\textbf{gCNR}} &
  0.884 &
  0.923 {\scriptsize (4$\%\uparrow$)}&
  0.795 {\scriptsize (10$\%\downarrow$)}&
  0.833 {\scriptsize (6$\%\downarrow$)}&
  0.912 {\scriptsize (3$\%\uparrow$)}&
  0.925 {\scriptsize (5$\%\uparrow$)}&
  0.907 {\scriptsize (3$\%\uparrow$)}&
  0.888 {\scriptsize (-)}&
  0.877 {\scriptsize (1$\%\downarrow$)}\\
 &
  \textit{\textbf{SNR}} &
  8.33 &
  12.82 {\scriptsize (54$\%\uparrow$)}&
  11.87 {\scriptsize (42$\%\uparrow$)}&
  11.47 {\scriptsize (38$\%\uparrow$)}&
  8.98 {\scriptsize (8$\%\uparrow$)}&
  9.76 {\scriptsize (17$\%\uparrow$)}&
  9.26 {\scriptsize (11$\%\uparrow$)}&
  8.45 {\scriptsize (1$\%\uparrow$)}&
  9.16 {\scriptsize (10$\%\uparrow$)}\\
 &
  \textit{\textbf{PSNR}} &
  17.22 &
  24.72 {\scriptsize (44$\%\uparrow$)}&
  22.53 {\scriptsize (31$\%\uparrow$)}&
  22.24 {\scriptsize (29$\%\uparrow$)}&
  19.30 {\scriptsize (12$\%\uparrow$)}&
  18.28 {\scriptsize (6$\%\uparrow$)}&
  19.83 {\scriptsize (15$\%\uparrow$)}&
  17.48 {\scriptsize (2$\%\uparrow$)}&
  19.54 {\scriptsize (13$\%\uparrow$)}\\
\multirow{-5}{*}{\textit{\textbf{Star}}} &
  \textit{\textbf{SSIM}} &
  0.261 &
  0.984 {\scriptsize (278$\%\uparrow$)}&
  0.635 {\scriptsize (144$\%\uparrow$)}&
  0.580 {\scriptsize (123$\%\uparrow$)}&
  0.390 {\scriptsize (50$\%\uparrow$)}&
  0.725 {\scriptsize (178$\%\uparrow$)}&
  0.387 {\scriptsize (48$\%\uparrow$)}&
  0.267 {\scriptsize (2$\%\uparrow$)}&
  0.371 {\scriptsize (42$\%\uparrow$)}\\ \midrule
 &
  \textit{\textbf{CNR}} &
  5.69 &
  7.83 {\scriptsize (38$\%\uparrow$)}&
  3.97 {\scriptsize (30$\%\downarrow$)}&
  4.57 {\scriptsize (20$\%\downarrow$)}&
  6.46 {\scriptsize (13$\%\uparrow$)}&
  6.48 {\scriptsize (14$\%\uparrow$)}&
  6.21 {\scriptsize (9$\%\uparrow$)}&
  6.11 {\scriptsize (7$\%\uparrow$)}&
  6.01 {\scriptsize (6$\%\uparrow$)}\\
 &
  \textit{\textbf{gCNR}} &
  0.811 &
  0.855 {\scriptsize (5$\%\uparrow$)}&
  0.718 {\scriptsize (11$\%\downarrow$)}&
  0.742 {\scriptsize (8$\%\downarrow$)}&
  0.841 {\scriptsize (4$\%\uparrow$)}&
  0.867 {\scriptsize (7$\%\uparrow$)}&
  0.829 {\scriptsize (2$\%\uparrow$)}&
  0.816 {\scriptsize (1$\%\uparrow$)}&
  0.813 {\scriptsize (-)}\\
 &
  \textit{\textbf{SNR}} &
  5.86 &
  10.82 {\scriptsize (85$\%\uparrow$)}&
  8.83 {\scriptsize (51$\%\uparrow$)}&
  8.38 {\scriptsize (43$\%\uparrow$)}&
  6.20 {\scriptsize (6$\%\uparrow$)}&
  6.58 {\scriptsize (12$\%\uparrow$)}&
  6.37 {\scriptsize (9$\%\uparrow$)}&
  5.95 {\scriptsize (2$\%\uparrow$)}&
  6.32 {\scriptsize (8$\%\uparrow$)}\\
 &
  \textit{\textbf{PSNR}} &
  15.15 &
  22.36 {\scriptsize (48$\%\uparrow$)}&
  19.26 {\scriptsize (27$\%\uparrow$)}&
  19.06 {\scriptsize (26$\%\uparrow$)}&
  16.61 {\scriptsize (10$\%\uparrow$)}&
  14.76 {\scriptsize ((3$\%\downarrow$)}&
  17.04 {\scriptsize (12$\%\uparrow$)}&
  15.36 {\scriptsize (1$\%\uparrow$)}&
  16.84 {\scriptsize (11$\%\uparrow$)}\\
\multirow{-5}{*}{\textit{\textbf{Circle}}} &
  \textit{\textbf{SSIM}} &
  0.235 &
  0.936 {\scriptsize (299$\%\uparrow$)}&
  0.539 {\scriptsize (130$\%\uparrow$)}&
  0.487 {\scriptsize (108$\%\uparrow$)}&
  0.330 {\scriptsize (41$\%\uparrow$)}&
  0.618 {\scriptsize (164$\%\uparrow$)}&
  0.323 {\scriptsize (38$\%\uparrow$)}&
  0.240 {\scriptsize (2$\%\uparrow$)}&
  0.307 {\scriptsize (31$\%\uparrow$)}\\ \bottomrule
\end{tabular}
}
\end{table*}

\subsubsection{Dual-domain constrain}

We formulate A2A as a dual-domain optimization problem by incorporating the beamformed IQ-to-B-mode image formation model. Consider a target view with IQ tensor $X_j\in\mathbb{C}^{N\times H\times W}$, let $X_{j}(\bm p)=r_{j}(\bm p)e^{\mathrm{j}\phi_{j}(\bm p)}$ denote the IQ signal at tensor location $\bm p$. The signal-domain reconstruction loss can be expressed as
\begin{equation}
\label{signal_loss}
\begin{aligned}
\mathcal{L}^{\mathrm{sig}}_{i\rightarrow j}
&=\frac{\left\|\widetilde{X}_{j\mid i}-X_j\right\|_F^2}{NHW}
=\frac{1}{NHW}\sum\Big[
(\widetilde{r}_{j\mid i}-r_{j})^2 \\
&\quad+
2\widetilde{r}_{j\mid i}r_{j}
\big(1-\cos(\widetilde{\phi}_{j\mid i}-\phi_{j})\big)
\Big]_{\bm p},
\end{aligned}
\end{equation}
where the first term penalizes amplitude errors and the second term penalizes the amplitude-weighted phase errors.

To evaluate the perceptual consistency in the B-mode images within the clinically displayed dynamic range, coherent compounding is first performed along the sub-aperture dimension:
\begin{equation}
\label{coherent_compounding}
\overline{X} = \frac{1}{N} \sum_{n=1}^{N} X[n,:,:]
\in\mathbb{C}^{H\times W}.
\end{equation}
Since equal-weight compounding is permutation invariant, all shuffled noisy samples share a consistent image-domain target. Envelope detection is then applied to $\overline{X}$ to get the compounded envelope $\overline{r}$. Defining a dynamic range of $D=50$ dB, the targeted B-mode image after log-compression and normalization is 
\begin{equation}
\label{bmode_db}
\mathcal{B} = \frac{\min \left(0, \max(-D, 20\log\overline{r} )\right) +D}{D} \in[0,1]^{H\times W}.
\end{equation}
The image-domain consistency loss is computed after this IQ-to-B-mode transformation:
\begin{equation}
\label{image_loss}
\mathcal{L}^{\mathrm{img}}_{i\rightarrow j} = \frac{1}{HW}
\left\| \mathcal{B}_{j\mid i} - \mathcal{B} \right\|_F^2.
\end{equation}

Collecting $M$ directional reconstructions, the complete dual-domain reconstruction loss is formulated as
\begin{equation}
\label{swap}
\begin{aligned}
\mathcal{L}_{\mathrm{rec}} = \frac{1}{M} \sum_{i=1}^M
\left(
\mathcal{L}^{\mathrm{sig}}_{i\rightarrow j} + \lambda_{\mathrm{img}}\mathcal{L}^{\mathrm{img}}_{i\rightarrow j}
\right),
\end{aligned}
\end{equation}
where $\lambda_{\mathrm{img}}$ balances the complementary IQ and B-mode fidelity.

The total objective of PSCD is formulated by combining the self-contrastive loss and the reconstruction loss:
\begin{equation}
\label{object}
\begin{aligned}
\min_{f_a, f_n, f_d} \mathbb{E}_{\mathcal{X}\sim X} \ (\mathcal{L}_{\mathrm{con}}+ \lambda_{\mathrm{rec}}\mathcal{L}_{\mathrm{rec}}),
\end{aligned}
\end{equation}
where $\lambda_{\mathrm{rec}}$ balances the effect of coherence-based disentanglement and A2A reconstruction.

Once the test-time training has converged, $f_d$ reconstructs the clean IQ tensor $Y$ by decoding the anatomy space exclusively from the original noisy tensor $X$ using Eq. \ref{hat}.

\section{Experiments}

\subsection{Simulation Setup}

\begin{figure}
\centering
\includegraphics[width=\columnwidth]{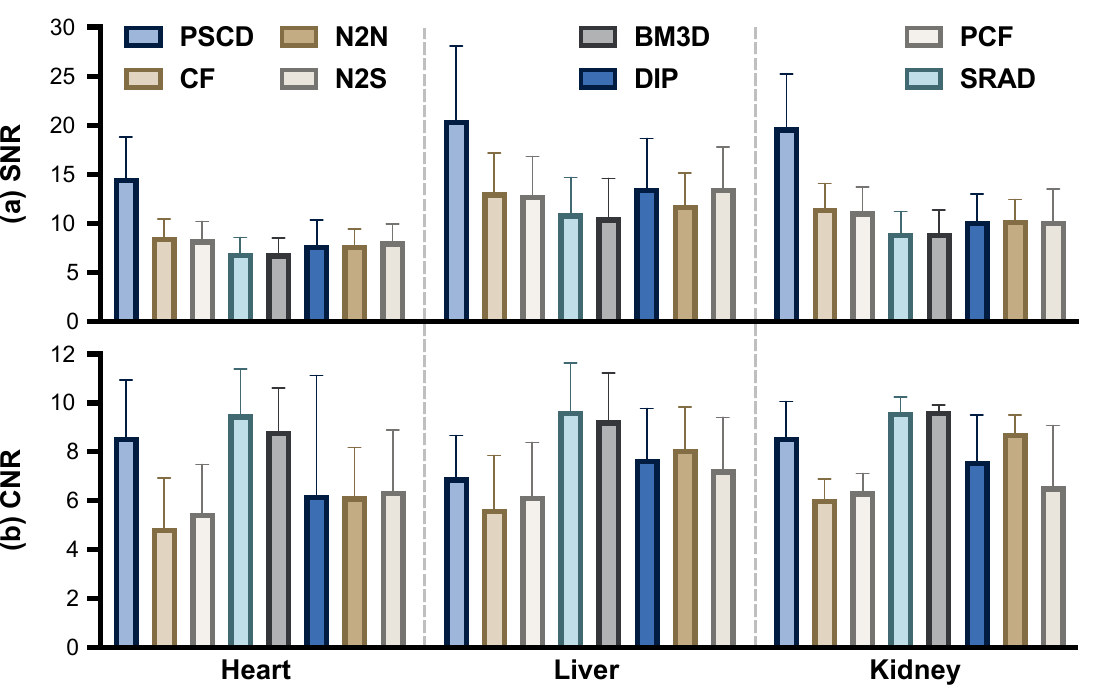}
\caption{The (a) SNR and (b) CNR among PSCD and seven comparison methods on \textit{in vivo} ultrasound images of different organs.}
\label{organ_hist}
\end{figure}

We used the k-Wave toolbox \cite{treeby2012modeling} to simulate SAU images in the P4-2 phased array configuration. As Table \ref{parameter}, the full aperture of 64 elements was divided into four and eight sub-apertures to assess performance under different imaging configurations. SNRs of 0, 10, 20, and 30 dB were simulated to evaluate robustness to various electronic noise levels. To consider targets of different complexities, we simulated circular and star-shaped inclusions whose acoustic properties were set with reference to those of the heart muscle. The medium parameters are summarized in Table \ref{medium}. The binary masks defining the inclusion shape and location are treated as structural references for PSNR and SSIM evaluation.

\begin{table*}
\centering
\caption{The SNR, CNR, and gCNR achieved by PSCD and comparison methods on different \textit{in vivo} organs and scan views.}
\label{invivo_metric}
\scalebox{0.85}{
\setlength{\tabcolsep}{7pt}
\begin{tabular}{@{}ccc|ccccccccc@{}}
\toprule
\textbf{Metrics} &
  \textbf{Organ} &
  \textbf{Scan View} &
  \textbf{Original} &
  \textbf{PSCD} &
  \textbf{CF} &
  \textbf{PCF} &
  \textbf{SRAD} &
  \textbf{BM3D} &
  \textbf{DIP} &
  \textbf{N2N} &
  \textbf{N2S} \\ \midrule
\multirow{8}{*}{\textbf{\textit{SNR}}} &
  \multirow{6}{*}{\textit{\textbf{Heart}}} &
  \textit{\textbf{A4C}} &
  8.27 &
  17.50 {\scriptsize (112$\%\uparrow$)}&
  9.20 {\scriptsize (11$\%\uparrow$)}&
  8.88 {\scriptsize (7$\%\uparrow$)}&
  7.07 {\scriptsize (14$\%\downarrow$)}&
  7.05 {\scriptsize (15$\%\downarrow$)}&
  7.66 {\scriptsize (7$\%\downarrow$)}&
  7.98 {\scriptsize (4$\%\downarrow$)}&
  7.14 {\scriptsize (14$\%\downarrow$)}\\
 &          & \textit{\textbf{A2C}}              &
 8.03  & 
 13.04 {\scriptsize (62$\%\uparrow$)}& 
 8.75  {\scriptsize (9$\%\uparrow$)}& 
 8.48  {\scriptsize (6$\%\uparrow$)}& 
 7.00  {\scriptsize (13$\%\downarrow$)}& 
 7.07  {\scriptsize (12$\%\downarrow$)}& 
 7.97  {\scriptsize (1$\%\downarrow$)}& 
 7.65  {\scriptsize (5$\%\downarrow$)}& 
 8.36  {\scriptsize (4$\%\uparrow$)}\\
 &          & \textit{\textbf{PSAX-MV}}          & 
 7.88  & 
 14.06 {\scriptsize (78$\%\uparrow$)}& 
 8.50  {\scriptsize (8$\%\uparrow$)}& 
 8.30  {\scriptsize (5$\%\uparrow$)}& 
 7.01  {\scriptsize (11$\%\downarrow$)}& 
 6.95  {\scriptsize (12$\%\downarrow$)}& 
 8.44  {\scriptsize (7$\%\uparrow$)}& 
 7.71  {\scriptsize (2$\%\downarrow$)}& 
 8.14  {\scriptsize (3$\%\uparrow$)}\\
 &          & \textit{\textbf{PSAX-Pap}}         &
 6.90  & 
 13.20 {\scriptsize (91$\%\uparrow$)}& 
 7.37  {\scriptsize (7$\%\uparrow$)}& 
 7.18  {\scriptsize (4$\%\uparrow$)}& 
 6.08  {\scriptsize (12$\%\downarrow$)}& 
 6.02  {\scriptsize (13$\%\downarrow$)}& 
 7.01  {\scriptsize (2$\%\uparrow$)}& 
 6.79  {\scriptsize (2$\%\downarrow$)}& 
 7.80  {\scriptsize (13$\%\uparrow$)}\\
 &          & \textit{\textbf{PSAX-Apex}}        & 
 8.32  & 
 13.82 {\scriptsize (66$\%\uparrow$)}& 
 8.90  {\scriptsize (7$\%\uparrow$)}& 
 8.67  {\scriptsize (4$\%\uparrow$)}& 
 7.32  {\scriptsize (12$\%\downarrow$)}& 
 7.26  {\scriptsize (13$\%\downarrow$)}& 
 8.04  {\scriptsize (3$\%\downarrow$)}& 
 8.18  {\scriptsize (2$\%\downarrow$)}& 
 8.74  {\scriptsize (5$\%\uparrow$)}\\
 &          & \textit{\textbf{PLAX}}             & 
 8.26  & 
 15.06 {\scriptsize (82$\%\uparrow$)}& 
 8.94  {\scriptsize (8$\%\uparrow$)}& 
 8.75  {\scriptsize (6$\%\uparrow$)}& 
 7.49  {\scriptsize (9$\%\downarrow$)}& 
 7.42  {\scriptsize (10$\%\downarrow$)}& 
 7.89  {\scriptsize (5$\%\downarrow$)}& 
 8.10  {\scriptsize (2$\%\downarrow$)}& 
 8.69  {\scriptsize (5$\%\uparrow$)}\\ \cmidrule(l){2-12} 
 & \multicolumn{2}{c|}{\textit{\textbf{Liver}}}  &
 12.99 & 
 24.45 {\scriptsize (88$\%\uparrow$)}& 
 13.80 {\scriptsize (6$\%\uparrow$)}& 
 13.60 {\scriptsize (5$\%\uparrow$)}& 
 11.79 {\scriptsize (9$\%\downarrow$)}& 
 11.47 {\scriptsize (12$\%\downarrow$)}& 
 14.42 {\scriptsize (11$\%\uparrow$)}& 
 12.43 {\scriptsize (4$\%\downarrow$)}& 
 14.37 {\scriptsize (11$\%\uparrow$)}\\
 & \multicolumn{2}{c|}{\textit{\textbf{Kidney}}} & 
 10.56 &
 19.78 {\scriptsize (87$\%\uparrow$)}& 
 11.58 {\scriptsize (10$\%\uparrow$)}& 
 11.20 {\scriptsize (6$\%\uparrow$)}& 
 8.97  {\scriptsize (15$\%\downarrow$)}& 
 8.97  {\scriptsize (15$\%\downarrow$)}& 
 10.23 {\scriptsize (3$\%\downarrow$)}& 
 10.33 {\scriptsize (2$\%\downarrow$)}& 
 10.20 {\scriptsize (3$\%\downarrow$)}\\ \midrule \specialrule{0em}{0.5pt}{0.5pt} \midrule
 \multirow{8}{*}{\textbf{\textit{CNR}}} &
  \multirow{6}{*}{\textit{\textbf{Heart}}} &
  \textit{\textbf{A4C}} &
  6.30 &
  9.01 {\scriptsize (43$\%\uparrow$)}&
  5.51 {\scriptsize (13$\%\downarrow$)}&
  5.93 {\scriptsize (6$\%\downarrow$)}&
  7.07 {\scriptsize (12$\%\uparrow$)}&
  9.34 {\scriptsize (48$\%\uparrow$)}&
  5.80 {\scriptsize (8$\%\downarrow$)}&
  6.30 {\scriptsize (-)}&
  4.71 {\scriptsize (25$\%\downarrow$)}\\
 &
   &
  \textit{\textbf{A2C}} &
  5.50 &
  6.84 {\scriptsize (24$\%\uparrow$)}&
  4.09 {\scriptsize (26$\%\downarrow$)}&
  4.71 {\scriptsize (14$\%\downarrow$)}&
  8.71 {\scriptsize (58$\%\uparrow$)}&
  8.43 {\scriptsize (53$\%\uparrow$)}&
  6.18 {\scriptsize (12$\%\uparrow$)}&
  5.46 {\scriptsize (1$\%\downarrow$)}&
  6.07 {\scriptsize (10$\%\uparrow$)}\\
 &
   &
  \textit{\textbf{PSAX-MV}} &
  5.62 &
  7.87 {\scriptsize (40$\%\uparrow$)}&
  4.47 {\scriptsize (20$\%\downarrow$)}&
  5.12 {\scriptsize (9$\%\downarrow$)}&
  9.37 {\scriptsize (67$\%\uparrow$)}&
  8.69 {\scriptsize (55$\%\uparrow$)}&
  7.03 {\scriptsize (25$\%\uparrow$)}&
  5.64 {\scriptsize (-)}&
  5.67 {\scriptsize (1$\%\uparrow$)}\\
 &
   &
  \textit{\textbf{PSAX-Pap}} &
  5.15 &
  7.44 {\scriptsize (44$\%\uparrow$)}&
  3.53 {\scriptsize (32$\%\downarrow$)}&
  4.28 {\scriptsize (17$\%\downarrow$)}&
  8.63 {\scriptsize (68$\%\uparrow$)}&
  7.75 {\scriptsize (50$\%\uparrow$)}&
  5.40 {\scriptsize (5$\%\uparrow$)}&
  5.17 {\scriptsize (-)}&
  6.78 {\scriptsize (32$\%\uparrow$)}\\
 &
   &
  \textit{\textbf{PSAX-Apex}} &
  7.19 &
  9.25 {\scriptsize (29$\%\uparrow$)}&
  5.72 {\scriptsize (20$\%\downarrow$)}&
  6.34 {\scriptsize (12$\%\downarrow$)}&
  10.08 {\scriptsize (40$\%\uparrow$)}&
  9.56 {\scriptsize (33$\%\uparrow$)}&
  7.80 {\scriptsize (8$\%\uparrow$)}&
  7.25 {\scriptsize (1$\%\uparrow$)}&
  8.04 {\scriptsize (12$\%\uparrow$)}\\
 &
   &
  \textit{\textbf{PLAX}} &
  6.50 &
  9.13 {\scriptsize (41$\%\uparrow$)}&
  5.31 {\scriptsize (18$\%\downarrow$)}&
  6.03 {\scriptsize (7$\%\downarrow$)}&
  10.04 {\scriptsize (55$\%\uparrow$)}&
  9.27 {\scriptsize (43$\%\uparrow$)}&
  5.81 {\scriptsize (11$\%\downarrow$)}&
  6.55 {\scriptsize (1$\%\uparrow$)}&
  6.99 {\scriptsize (12$\%\uparrow$)}\\ \cmidrule(l){2-12} 
 &
  \multicolumn{2}{c|}{\textit{\textbf{Liver}}} &
  6.60 &
  6.94 {\scriptsize (5$\%\uparrow$)}&
  5.64 {\scriptsize (14$\%\downarrow$)}&
  6.19 {\scriptsize (6$\%\downarrow$)}&
  9.65 {\scriptsize (46$\%\uparrow$)}&
  9.28 {\scriptsize (41$\%\uparrow$)}&
  7.83 {\scriptsize (19$\%\uparrow$)}&
  8.30 {\scriptsize (26$\%\uparrow$)}&
  7.44 {\scriptsize (13$\%\uparrow$)}\\
 &
  \multicolumn{2}{c|}{\textit{\textbf{Kidney}}} &
  6.92 &
  8.58 {\scriptsize (24$\%\uparrow$)}&
  6.07 {\scriptsize (12$\%\downarrow$)}&
  6.38 {\scriptsize (8$\%\downarrow$)}&
  9.62 {\scriptsize (39$\%\uparrow$)}&
  9.66 {\scriptsize (40$\%\uparrow$)}&
  7.61 {\scriptsize (10$\%\uparrow$)}&
  8.76 {\scriptsize (27$\%\uparrow$)}&
  6.57 {\scriptsize (5$\%\downarrow$)}\\ \midrule \specialrule{0em}{0.5pt}{0.5pt} \midrule
  \multirow{8}{*}{\textbf{\textit{gCNR}}} &
  \multirow{6}{*}{\textit{\textbf{Heart}}} &
  \textit{\textbf{A4C}} &
  0.871 &
  0.917 {\scriptsize (5$\%\uparrow$)}&
  0.812 {\scriptsize (7$\%\downarrow$)}&
  0.832 {\scriptsize (4$\%\downarrow$)}&
  0.974 {\scriptsize (12$\%\uparrow$)}&
  0.971 {\scriptsize (11$\%\uparrow$)}&
  0.835 {\scriptsize (4$\%\downarrow$)}&
  0.868 {-)}&
  0.775 {\scriptsize (11$\%\downarrow$)}\\
 &
   &
  \textit{\textbf{A2C}} &
  0.825 &
  0.848 {\scriptsize (3$\%\uparrow$)}&
  0.750 {\scriptsize (9$\%\downarrow$)}&
  0.768 {\scriptsize (7$\%\downarrow$)}&
  0.948 {\scriptsize (15$\%\uparrow$)}&
  0.944 {\scriptsize (14$\%\uparrow$)}&
  0.836 {\scriptsize (1$\%\uparrow$)}&
  0.821 {\scriptsize (-)}&
  0.844 {\scriptsize (2$\%\uparrow$)}\\
 &
   &
  \textit{\textbf{PSAX-MV}} &
  0.828 &
  0.879 {\scriptsize (6$\%\uparrow$)}&
  0.770 {\scriptsize (7$\%\downarrow$)}&
  0.796 {\scriptsize (4$\%\downarrow$)}&
  0.949 {\scriptsize (15$\%\uparrow$)}&
  0.945 {\scriptsize (14$\%\uparrow$)}&
  0.861 {\scriptsize (1$\%\uparrow$)}&
  0.825 {\scriptsize (-)}&
  0.815 {\scriptsize (2$\%\downarrow$)}\\
 &
   &
  \textit{\textbf{PSAX-Pap}} &
  0.808 &
  0.864 {\scriptsize (7$\%\uparrow$)}&
  0.721 {\scriptsize (11$\%\downarrow$)}&
  0.750 {\scriptsize (7$\%\downarrow$)}&
  0.941 {\scriptsize (16$\%\uparrow$)}&
  0.915 {\scriptsize (13$\%\uparrow$)}&
  0.816 {\scriptsize (1$\%\uparrow$)}&
  0.804 {\scriptsize (-)}&
  0.873 {\scriptsize (8$\%\uparrow$)}\\
 &
   &
  \textit{\textbf{PSAX-Apex}} &
  0.880 &
  0.902 {\scriptsize (2$\%\uparrow$)}&
  0.798 {\scriptsize (9$\%\downarrow$)}&
  0.833 {\scriptsize (5$\%\downarrow$)}&
  0.964 {\scriptsize (10$\%\uparrow$)}&
  0.953 {\scriptsize (8$\%\uparrow$)}&
  0.902 {\scriptsize (3$\%\uparrow$)}&
  0.877 {\scriptsize (-)}&
  0.910 {\scriptsize (3$\%\uparrow$)}\\
 &
   &
  \textit{\textbf{PLAX}} &
  0.864 &
  0.912 {\scriptsize (6$\%\uparrow$)}&
  0.808 {\scriptsize (6$\%\downarrow$)}&
  0.835 {\scriptsize (3$\%\downarrow$)}&
  0.965 {\scriptsize (12$\%\uparrow$)}&
  0.957 {\scriptsize (11$\%\uparrow$)}&
  0.825 {\scriptsize (5$\%\downarrow$)}&
  0.861 {\scriptsize (-)}&
  0.870 {\scriptsize (1$\%\uparrow$)}\\ \cmidrule(l){2-12} 
 &
  \multicolumn{2}{c|}{\textit{\textbf{Liver}}} &
  0.877 &
  0.896 {\scriptsize (2$\%\uparrow$)}&
  0.838 {\scriptsize (4$\%\downarrow$)}&
  0.857 {\scriptsize (2$\%\downarrow$)}&
  0.974 {\scriptsize (11$\%\uparrow$)}&
  0.983 {\scriptsize (12$\%\uparrow$)}&
  0.922 {\scriptsize (5$\%\uparrow$)}&
  0.944 {\scriptsize (8$\%\uparrow$)}&
  0.906 {\scriptsize (3$\%\uparrow$)}\\
 &
  \multicolumn{2}{c|}{\textit{\textbf{Kidney}}} &
  0.881 &
  0.918 {\scriptsize (4$\%\uparrow$)}&
  0.836 {\scriptsize (5$\%\downarrow$)}&
  0.846 {\scriptsize (4$\%\downarrow$)}&
  0.974 {\scriptsize (11$\%\uparrow$)}&
  0.995 {\scriptsize (13$\%\uparrow$)}&
  0.879 {\scriptsize (-)}&
  0.941 {\scriptsize (7$\%\uparrow$)}&
  0.836 {\scriptsize (5$\%\downarrow$)}\\ \bottomrule
\end{tabular}
}
\end{table*}

\begin{figure*}
\centering
\includegraphics[width=\textwidth]{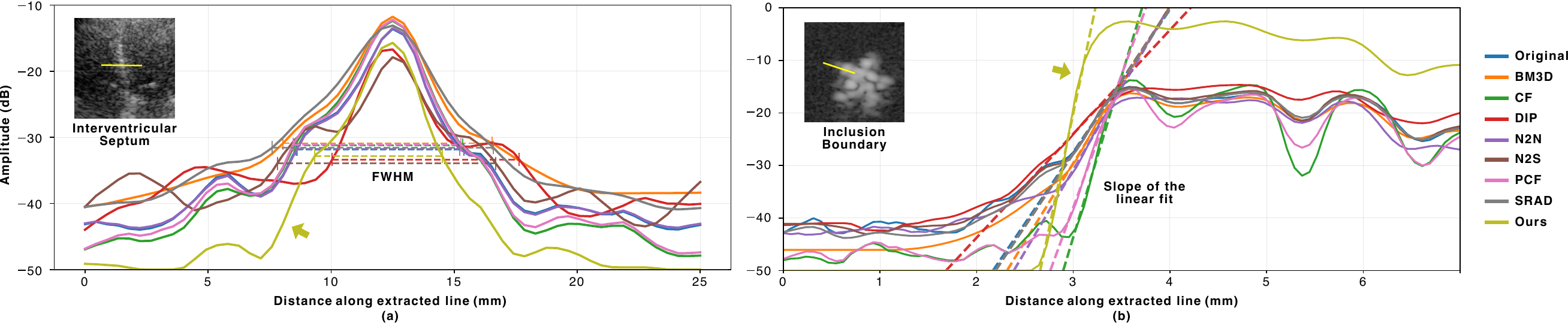}
\caption{\color{black} (a) Amplitude profiles across the interventricular septum, where the FWHM (dashed lines) measures the capability to resolve fine anatomical structures. (b) Profiles across an inclusion boundary, where the slope of the linear fit (dashed lines) quantifies boundary sharpness. All profiles are measured under the same dynamic range.}
\label{profile}
\end{figure*}

\subsection{\textit{In Vivo} Data Acquisition}

We collected a large \textit{in vivo} human SAU dataset consisting of echocardiographic and abdominal images. The cohort included 74 healthy subjects and 21 patients with hypertension and/or diabetes. Two sonographers performed the scans. Raw IQ signals were acquired using a Vantage 256 system (Verasonics, Kirkland, WA, USA) equipped with a P4-2 phased array \footnote{Human experiment protocols were approved by the Institutional Review Board of The University of Hong Kong (UW19-043, EA250292).}. The detailed imaging parameters are summarized in Table \ref{parameter}. Individually beamformed frame from each sub-aperture measurement was scan-converged and resized to $512\times512$ pixels.

In this SAU dataset, the echocardiographic part includes six standard views (Fig. \ref{domain}(a)), including apical four-chamber (A4C) view, apical two-chamber (A2C) view, parasternal short-axis views at the mitral valve (PSAX-MV), papillary muscle (PSAX-Pap), and apical (PSAX-Apex) levels, and the parasternal long-axis (PLAX) view. There were 16.4k, 16k, 13.6k, 16.48k, 16.24k, and 15.59k images for these views, respectively. The abdominal part includes 4.8k images of the liver and kidney. This diverse dataset supports evaluations under domain shifts across scan views, organs, and subject conditions.

\subsection{Implementation Details}
The PSCD \footnote{The code and data will be publicly available on \url{https://github.com/JustinJZhang/PSCD/tree/main}.} model was implemented in Python 3.10, PyTorch 2.1.2, CUDA 11.8, and executed on an NVIDIA RTX 3090 GPU using $<$6029 MiB of memory. 
In all experiments, we set $\lambda_{\mathrm{img}}=1$, $\lambda_{\mathrm{rec}}=1$, $\tau=1$, and $M=2$ for simplification.

Here we consider each SAU acquisition as an individual testing domain. The entire framework was optimized on the first frame of an SAU video at test time using an Adam optimizer with $L_2$ regularization. Early stopping is triggered by a 10-iteration plateau of $\mathcal{L}_{con}$. This single-shot optimization phase took 62.8 s on average. Then, PSCD can infer subsequent video frames at a speed of 0.04 s per frame.

\subsection{Comparison Methods}
We compared the image quality of our PSCD with seven representative ultrasound denoising methods described in related works (section \ref{related}). They include three types of techniques: (1) model-based SRAD \cite{yu2002speckle} and BM3D \cite{dabov2007image}; (2) coherence-based CF \cite{hollman1999coherence} and PCF \cite{camacho2009phase}; (3) learning-based DIP \cite{ulyanov2018deep}, N2N \cite{lehtinen2018noise2noise}, and N2S \cite{batson2019noise2self}. Similar to PSCD, DIP and N2S were trained at test time on each testing image. N2N was pretrained on 70\% of subjects and tested on the remaining 30\% of the subjects. Denoising performance was quantified by comprehensive metrics of CNR, gCNR, SNR, PSNR, and SSIM on diverse image domains illustrated in Fig. \ref{domain} (a).

\section{Results}

\subsection{Simulation Results}
\subsubsection{Noise level adaptation}
Table \ref{simu_metric} summarizes the results of simulation data under four electronic noise levels. When the original images suffer from severe electronic noise, i.e., at 0 dB SNR as shown in Fig. \ref{simu_compare}(a), our PSCD improved CNR by 41.6$\%$, gCNR by 6$\%$, SNR by 111.8$\%$, PSNR by 93.1$\%$, and SSIM by about fourfold. When the original images are dominated by speckle noise, i.e., at an SNR of 30 dB in Fig. \ref{simu_compare}(b), our PSCD improved CNR by 37.5$\%$, gCNR by 5.6$\%$, SNR by 54.3$\%$, PSNR by 31.7$\%$, and SSIM by about twofold. At SNRs of 10 dB and 20 dB, the PSCD also shows consistent performance and surpasses all comparison methods. 

Fig. \ref{simu_compare} shows that at 0 dB, the SRAD and BM3D still yielded noisy backgrounds, DIP suffered from random spots, and N2N generated radial textures. At 30 dB, CF, PCF, SRAD, DIP, and N2S could not effectively reduce strong speckles in the background. These differences highlighted the domain gap of these methods when encountering various noise levels. However, our PSCD produced a clean background and was robust to different noise levels.

\subsubsection{Inclusion geometry adaptation}
Table \ref{simu_geo} summarizes the denoising performance across the simulated star and circle geometries. PSCD achieved the highest CNR, SNR, PSNR, and SSIM against seven comparison methods, demonstrating robust adaptation to both simple and complex targets. While coherence-based methods produced higher SNR but lower CNR, model-based methods exhibit the opposite trend.

\subsubsection{Sub-apertures adaptation}
Fig. \ref{simu_radar} shows radar plots for four and eight sub-apertures. PSCD improved CNR by 42.1$\%$ and 50.5$\%$, gCNR by 5.8$\%$ and 5.9$\%$, SNR by 98.1$\%$ and 100.5$\%$, PSNR by 84.8$\%$ and 96.1$\%$, and SSIM both four times using four and eight apertures, respectively. Those results confirmed that PSCD could adapt to flexible system configurations.

\subsubsection{Boundary sharpness}
Fig. \ref{profile} (b) shows the amplitude profiles across an inclusion boundary, where the ideal profile should be a step function. The boundary sharpness was measured by the slope of the linear fit within each step region. PSCD outperformed all other methods with the highest slope, which was 3.2 times that of the slope in the original image.

\subsection{\textit{In Vivo} Results}
\subsubsection{Scan view adaptation}
PSCD consistently improved SNRs in six standard echocardiographic views. As listed in Table \ref{invivo_metric}, the average SNR of PSCD was 14.44 dB, with 81.8$\%$ higher than the original images. The box plots in Fig. \ref{echo_box} (a-f) illustrate the remarkably higher SNR distributions of PSCD than the comparison methods. PSCD also exhibited consistent improvements in image contrast in six views. PSCD's average CNR and gCNR were 8.25 dB and 0.887, which were 36.6$\%$ and 4.8$\%$ better than the original images. The CNR box plots in Fig. \ref{echo_box}(g-i) show that PSCD manifested relatively higher contrast than other methods. 

Fig. \ref{invivo_compare} (a-c) visualizes the denoising results of A4C, PSAX, and PLAX views. Model-based methods blurred the entire structure and elevated the signal intensity in the heart chambers. Learning-based methods showed some effectiveness in reducing Gaussian noise but failed to mitigate strong side lobes and speckles. PSCD significantly suppressed noise in the cardiac chambers while preserving structural details of the myocardium. Especially in Fig. \ref{invivo_compare}(a), PSCD eliminated the side lobes around the atrial and ventricular walls, thereby revealing sharp anatomical boundaries. Electronic noise in deeper regions was also removed by PSCD (Fig. \ref{invivo_compare}(b-c)).

\begin{table}
\centering
\caption{FWHM and boundary-slope measurements from the amplitude profiles shown in Fig.~\ref{profile}.}
\label{profile_metric}
\begin{tabular*}{0.8\columnwidth}
{@{\extracolsep{\fill}}cccc@{}}
\toprule
\multicolumn{2}{c}{}
& \makecell{\textbf{FWHM} ($\downarrow$)}
& \makecell{\textbf{Boundary slope} ($\uparrow$)} \\
\midrule
\multicolumn{2}{c}{Original}
& 6.81 mm
& 27.14 dB/mm \\
\midrule
\multirow{8}{*}{\makecell{Comparison\\Method}}
& CF
& 6.83 mm
& 61.31 dB/mm \\
& PCF
& 6.83 mm
& 50.53 dB/mm \\
& SRAD
& 8.98 mm
& 27.63 dB/mm \\
& BM3D
& 8.49 mm
& 29.83 dB/mm \\
& DIP
& 7.60 mm
& 19.76 dB/mm \\
& N2N
& 6.61 mm
& 31.20 dB/mm \\
& N2S
& 8.88 mm
& 27.71 dB/mm \\
& PSCD
& \textbf{5.18 mm}
& \textbf{87.56 dB/mm} \\
\bottomrule
\end{tabular*}
\end{table}

\begin{figure}
\centering
\includegraphics[width=\columnwidth]{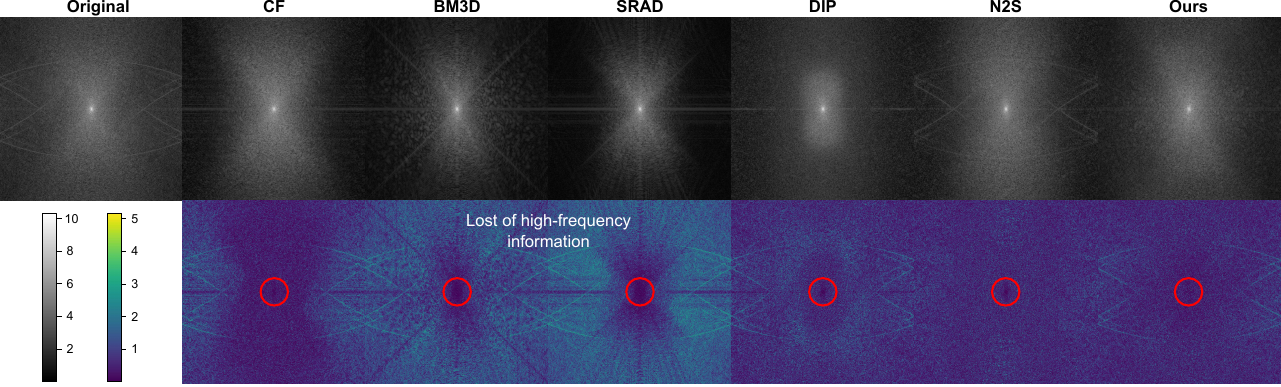}
\caption{\color{black} The spatial-frequency spectrum (top) of the \textit{in vivo} denoised results and their frequency difference (bottom) from the original one.}
\label{frequency}
\end{figure}

\begin{figure}
\centering
\includegraphics[width=\columnwidth]{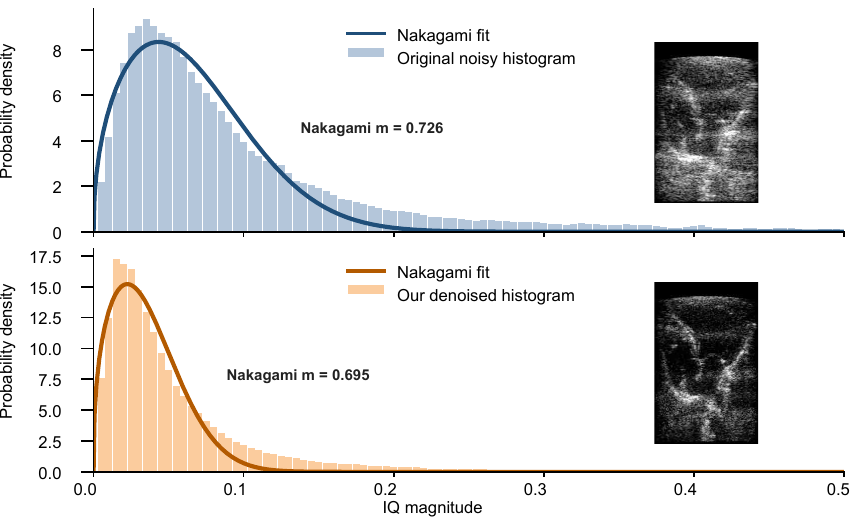}
\caption{\color{black} Speckle statistics of the original (top) and our reconstructed (bottom) IQ magnitude of an echocardiogram showing the \textit{in vivo} human heart.}
\label{speckle}
\end{figure}

\subsubsection{Imaging organ adaptation}

In addition to echocardiograms, Table \ref{invivo_metric} summarizes SNR, CNR, and gCNR of the \textit{in vivo} liver and kidney results. PSCD improved SNR, CNR, and gCNR on average by 87.8$\%$, 14.9$\%$, and 3.1$\%$. Fig. \ref{invivo_compare} (d-e) shows that the middle and left hepatic veins (MHV, LHV), as well as the internal structure in the liver ultrasound images, became clearer when subjected to PSCD denoising. PSCD reduced both electronic and speckle noise while preserving the signal intensity and detailed texture. SRAD and BM3D blurred the tissue texture (Fig. \ref{invivo_compare} (e)). Learning-based methods still struggle to remove noise in low-SNR environments. Fig. \ref{organ_hist} summarizes the cross-organ results, in which PSCD maintains a performance advantage over the comparison methods for \textit{in vivo} heart, liver, and kidney images.

\subsubsection{Image detail preservation}
\textcolor{black}{Fig. \ref{profile}(a) illustrates the amplitude profiles traversing the interventricular septum. We measured the Full Width at Half Maximum (FWHM) of this \textit{in vivo} line target to quantitatively evaluate the preservation of fine anatomical details. PSCD achieved the narrowest FWHM,} \textcolor{black}{demonstrating a 23.94\% improvement compared to the noisy input. BM3D, SRAD, and N2S yielded FWHM values exceeding 8 mm, compromising the delineation of fine structures.}

\textcolor{black}{Fig. \ref{frequency} compares the spatial frequency spectra of the noisy input and the denoised images. PSCD effectively preserved high-frequency details, indicated by low differences in the high-frequency band outside the red circle (15\% threshold). Conversely, BM3D and SRAD sacrificed considerable high-frequency information, leading to over-smoothed results.}

\subsubsection{Speckle statistic}
\textcolor{black}{Fig. \ref{speckle} compares the speckle statistic between the original and our reconstructed images. In the chamber regions, the grainy speckle patterns were significantly suppressed. In the myocardial tissue region, PSCD slightly altered the characteristic Nakagami distribution. The resulting shape parameter $m$ approximated that of the original image ($\triangle m=0.031$), demonstrating that PSCD suppressed background noise without sacrificing the informative speckle textures.}

\begin{figure*}
\centering
\includegraphics[width=\textwidth]{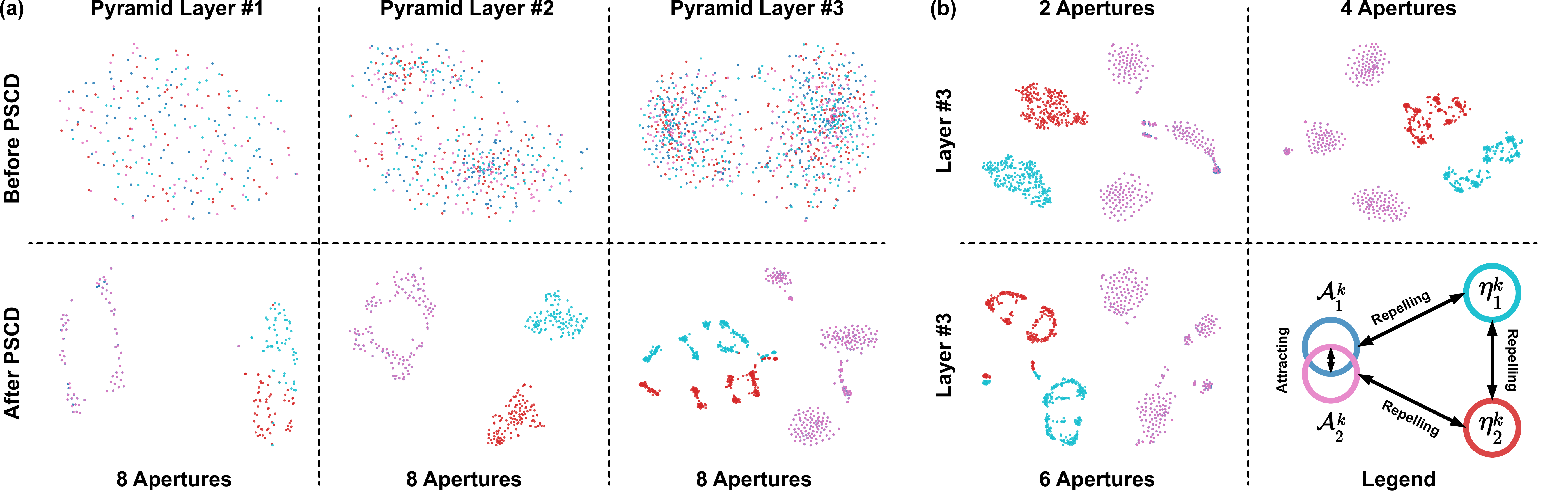}
\caption{(a) The pyramid feature distributions visualized by t-SNE using N=8 sub-apertures. (b) The feature distributions learned from N=2, 4, and 6 sub-apertures in the deepest layer. $\mathcal{A}_{1}^k, \mathcal{A}_{2}^k, \eta_{1}^k, \eta_{2}^k$ are coded in blue, pink, cyan, and red, respectively.}
\label{tSNE}
\end{figure*}

\begin{figure}
\centering
\includegraphics[width=\columnwidth]{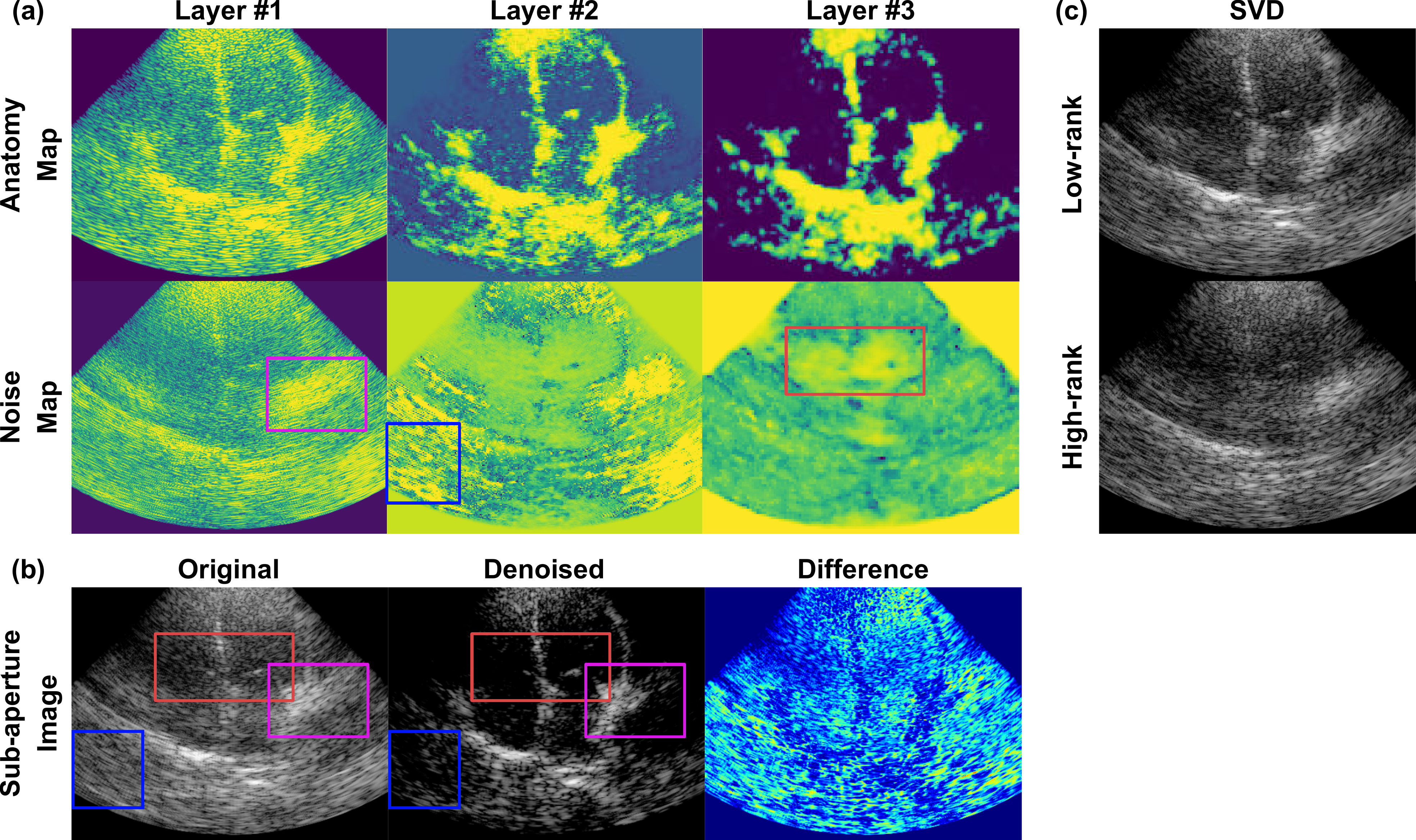}
\caption{(a) Visualization of PSCD disentanglement: feature maps extracted by $f_a$ and $f_n$. (b) The original and denoised result of one sub-aperture. (c) Comparison with SVD disentanglement.}
\label{feature}
\end{figure}

\section{Discussion}

The proposed SAU IQ signal denoising pipeline tightly couples dual encoders and one decoder in the PSCD framework and jointly optimizes them through the A2A TTT strategy. Standard ablation studies that remove either indispensable component are infeasible for this study. Yet we conducted a comprehensive analysis to verify the behavior and effectiveness of each component.

\subsection{Self-contrastive Disentanglement of Single-shot IQ Data}

The convergence of $\mathcal{L}_{\mathrm{con}}$ produces two feature spaces: an anatomy space featuring coherent components and a noise space featuring incoherent components. We visualized feature distributions via t-distributed stochastic neighbor embedding (t-SNE). 
As Fig. \ref{tSNE}(a) shows, the latent space before PSCD exhibited a random distribution where anatomy and noise features were intermingled. After PSCD, the features in each layer aggregated into three clusters: (1) an anatomy cluster with overlapping $\mathcal{A}_1^k$ and $\mathcal{A}_2^k$, representing shared information among sub-apertures; (2) a noise cluster with $\eta_1^k$ for unique noise components in shuffled sub-apertures $X_1$; (3) another noise cluster with $\eta_2^k$ unique noise components in shuffled sub-apertures $X_2$. The aggregation of anatomical features and the isolation of two noise clusters verify that PSCD effectively extracts anatomical coherence and noise incoherence from single-shot imaging data. As highlighted in the legend, the attraction between $(\mathcal{A}_1^k, \mathcal{A}_2^k)$ is optimized by Eq. \ref{max_a12}, while the repulsion between $(\eta_1^k, \eta_2^k)$ and between $(\mathcal{A}_{1/2}^k, \eta_{1/2}^k)$ are optimized by Eq. \ref{min_n12} and Eq. \ref{min_an}, respectively.

A larger aperture number increases incoherent noise variability due to aperture location-dependent side lobes and random electronic noise. When the aperture number rises from 2 to 6, PSCD learns increasingly sophisticated noise feature distributions (Fig. \ref{tSNE}(b)). This demonstrates that PSCD aligns with physical acoustic principles and is able to disentangle coherent and incoherent information regardless of the imaging configuration.

\subsection{Pyramid Latent Space for Dense Contrastive Learning}
Unlike conventional contrastive learning that merely manipulates the deepest features after a projection head \cite{oord2018representation}, the pyramid structure of our contrastive learning space enables multi-scale dense representations of coherent anatomy and incoherent noise. This is critical for ultrasound imaging since three noise types manifest at distinct spatial scales. Three layers of $f_n$ hierarchically extract side lobes, speckle patterns, and electronic noise as shown in Fig. \ref{feature}(a). These noises are thus comprehensively reduced via the clean IQ reconstruction, as shown by the comparison regions in Fig. \ref{feature}(b). The pyramid space of $f_a$ also facilitates a coarse-to-fine delineation of coherent anatomical information, preserving both macro structures and micro details in the clean output.

Fig. \ref{tSNE}(a) further reveals the hierarchical behaviors across three pyramid layers of $f_a$ and $f_n$. In the shallowest layer, $\eta_1^k$ and $\eta_2^k$ clusters were in close proximity, indicating similar low-level noise (e.g., electronic noise) across different sub-apertures. The deepest layer exhibits a more intricate distribution that splits into multiple subgroups, suggesting that high-level noise is sensitive to different sub-aperture locations (e.g., side lobes at different tilt angles).

Singular Value Decomposition (SVD) relies on linear transformations and manual threshold tuning, whereas PSCD transcends its low-rank prior in an adaptive, nonlinear, and multi-scale way. In Fig. \ref{feature}(a,c), the SVD low-rank component remains heavily contaminated by noise, resembling PSCD's anatomy map at the shallowest layer. Nevertheless, PSCD progressively captures clear cardiac structures in the deeper layers and can therefore robustly handle multiple scales of anatomical details.

\subsection{Aperture-to-Aperture Test-Time Training}
\begin{figure}
\centering
\includegraphics[width=\columnwidth]{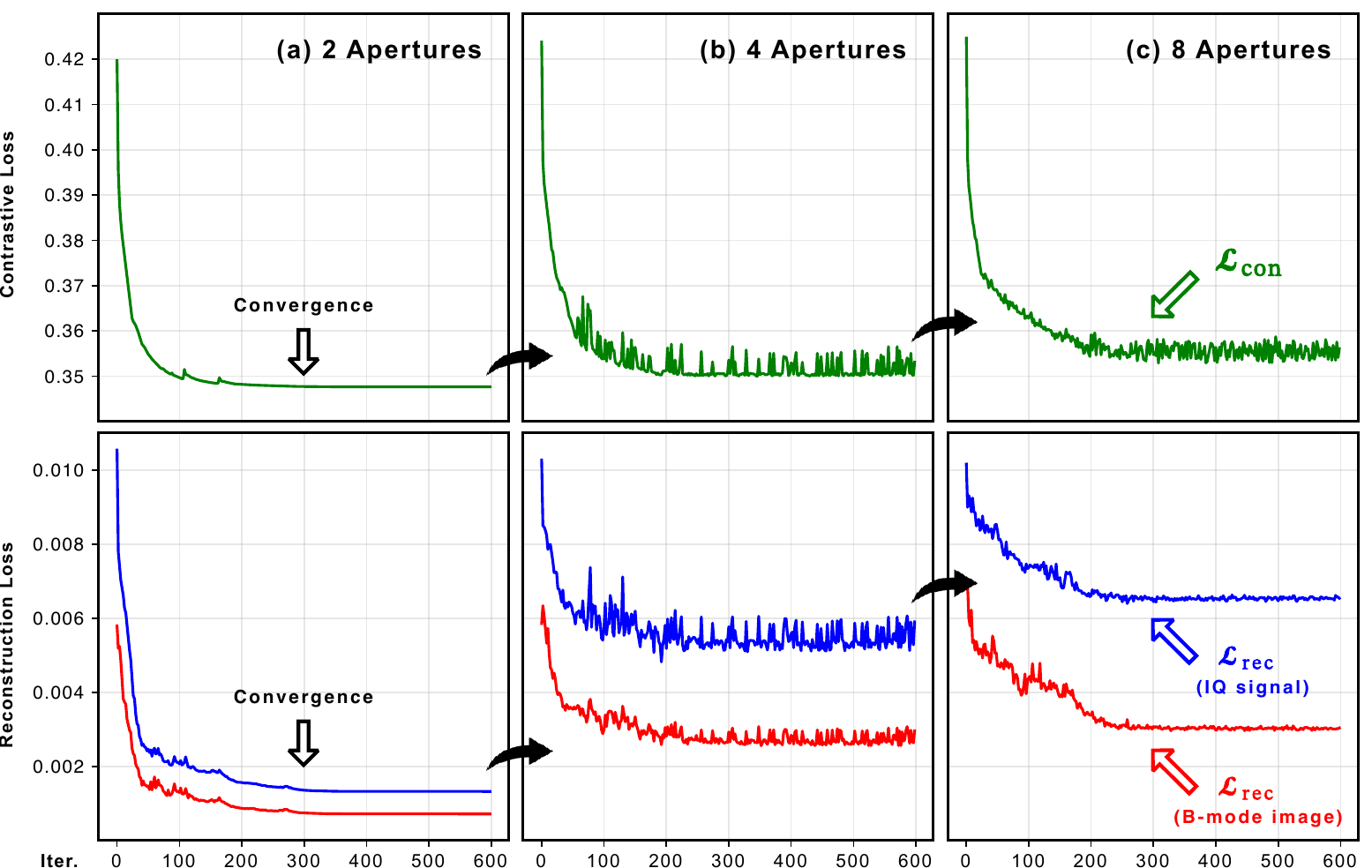}
\caption{The $\mathcal{L}_{con}$ and $\mathcal{L}_{swap}$ monitored with our test-time training process using (a) 2, (b) 4, and (c) 8 apertures.}
\label{loss}
\end{figure}

The whole PSCD model features a pretraining-free test-time training (TTT) using an Aperture-to-Aperture (A2A) strategy. DIP \cite{ulyanov2018deep}, likewise free from pretraining, leverages the implicit prior of CNNs that learn salient signals before learning noise. However, diagnostic details are not necessarily salient in SAU images, as useful signals and noise are highly coupled. PSCD generalizes the cross-aperture coherence theory in the raw IQ signal domain and demonstrates that the coherent and incoherent components are separable in a high-dimensional latent space. The convergence of this single-shot optimization is verified in Fig. \ref{loss} without the risk of overtraining. 

In practice, increasing the aperture number with a fixed total element count reduces the aperture size and acoustic energy for each sub-aperture transmission. This can degrade the initial SNR and lower the coherence among the received sub-aperture IQ signals. This system limitation is reflected in Fig. \ref{loss}, where a higher aperture number corresponds to convergence at elevated levels of both $\mathcal{L}_{\mathrm{con}}$ and $\mathcal{L}_{\mathrm{rec}}$.

The proposed A2A TTT method can be regarded as a generalized version of the N2N paradigm \cite{lehtinen2018noise2noise}, but is not bounded by the fixed pretraining domain. The random reconstruction from one noisy sample to another provides self-supervision ($\mathcal{L}_{\mathrm{rec}}$) on the fidelity of output IQ signals from $f_d$ using any encoded feature space. The clean IQ signal is then reconstructed exclusively from the anatomy space. The clean IQ reconstruction achieves noise reduction by discarding the noise space. Although random permutation on the sub-aperture dimension does not create new noise realizations or independence between noisy samples, our denoising effect arises from the coherence-regularized IQ feature representation and architectural inductive biases. Therefore, our framework relaxes the i.i.d. condition of N2N and can address the aperture-dependent signal degradation. 

\section{Conclusion}

This study proposes an adaptive SAU denoising framework that incorporates the pyramid self-contrastive disentanglement (PSCD) on the IQ signals guided by the cross-aperture spatial coherence prior. Our PSCD is pretraining-free and self-supervised by an Aperture-to-Aperture (A2A) test-time training strategy using only measured IQ data from single-shot SAU imaging, which thereby fundamentally addresses the challenges of domain shift and data scarcity. Simulation and \textit{in vivo} evaluations demonstrate that the denoised IQ signals boost B-mode image quality from the perspectives of SNR, image contrast, resolution, boundary sharpness, and structural detail preservation. In the future, this framework can be naturally generalized to other ultrasound transmit-receive modes that involve multiple noisy measurements. It holds significant promise for more precise anatomical visualization and more reliable clinical ultrasound assessment.

\bibliographystyle{IEEEtran}
\bibliography{ref}

\end{document}